\documentclass{article}

\usepackage[preprint]{corl_2026}

\usepackage[utf8]{inputenc}
\usepackage[T1]{fontenc}
\usepackage{hyperref}
\usepackage{url}
\usepackage{microtype}
\usepackage{graphicx}
\usepackage{amsmath,amssymb,amsthm}
\usepackage{booktabs}
\usepackage{multirow}
\usepackage{array}
\usepackage{xcolor}
\usepackage{enumitem}
\usepackage{subcaption}
\usepackage{wrapfig}
\usepackage{adjustbox}
\usepackage{makecell}
\usepackage{bm}
\usepackage{pifont}  % for \ding symbols in tables

\newif\ifdraft
\draftfalse   % flip to \draftfalse for camera-ready (hides all markers)

\definecolor{thesis3D}{HTML}{D2691E}
\definecolor{thesisCO}{HTML}{B8860B}
\definecolor{thesisMO}{HTML}{2E8B57}
\definecolor{thesisAU}{HTML}{6A5ACD}

\usepackage{xspace}
\newcommand{\egoinf}{\textsc{EgoInfinity}\xspace}

\newcommand{\actiondataset}{Action100M\xspace}
\newcommand{\mano}{\textsc{MANO}\xspace}
\newcommand{\moge}{\textsc{MoGe-2}\xspace}
\newcommand{\wilor}{\textsc{WiLoR}\xspace}
\newcommand{\samtwo}{\textsc{SAM-2}\xspace}
\newcommand{\samthree}{\textsc{SAM-3}\xspace}
\newcommand{\samthreed}{\textsc{SAM-3D}\xspace}

\newcommand{\geocalib}{\textsc{GeoCalib}\xspace}

\newcommand{\foundationposepp}{\textsc{FoundationPose++}\xspace}
\newcommand{\flowthreer}{\textsc{Flow3r}\xspace}
\newcommand{\memfof}{\textsc{MEMFOF}\xspace}

\newcommand{\Rmat}{\mathbf{R}}
\newcommand{\tvec}{\mathbf{t}}

\newcommand{\thand}{\boldsymbol{\theta}^{h}}
\newcommand{\bhand}{\boldsymbol{\beta}^{h}}

\newcommand{\pose}{\mathbf{p}}

\newcommand{\gravity}{\mathbf{g}}

\setlist[itemize]{leftmargin=*,topsep=2pt,itemsep=1pt,parsep=0pt}
\setlist[enumerate]{leftmargin=*,topsep=2pt,itemsep=1pt,parsep=0pt}

\newcommand{\figref}[1]{Fig.~\ref{#1}}
\newcommand{\tabref}[1]{Tab.~\ref{#1}}
\newcommand{\secref}[1]{Sec.~\ref{#1}}

\graphicspath{{figures/}}

\title{EgoInfinity: A Web-Scale 4D Hand-Object Interaction Data Engine for Any-View Robot Retargeting and Video-to-Action Robot Learning}

\author{
  Gaotian Wang \\
  Rice University \\
  \texttt{gw23@rice.edu} \\
  \And
  Kejia Ren \\
  Rice University \\
  \texttt{kr43@rice.edu} \\
  \And
  Andrew Morgan \\
  Robotics and AI Institute \\
  \texttt{andy@rai-inst.com} \\
  \AND
  Yiting Chen \\
  Rice University \\
  \texttt{yc203@rice.edu} \\
  \And
  Howard H. Qian \\
  Rice University \\
  \texttt{hhq1@rice.edu} \\
  \And
  Podshara Chanrungmaneekul \\
  Rice University \\
  \texttt{pc45@rice.edu} \\
  \AND
  Kaiyu Hang \\
  Rice University \\
  \texttt{kaiyu.hang@rice.edu} \\
}

\begin{document}
\newgeometry{textwidth=6.5in, textheight=9in, top=1in, headheight=12pt, headsep=25pt, footskip=30pt}
\maketitle

% =====================================================================
% Abstract
% =====================================================================
\vspace{-20pt}
\begin{center}
  \includegraphics[width=\linewidth]{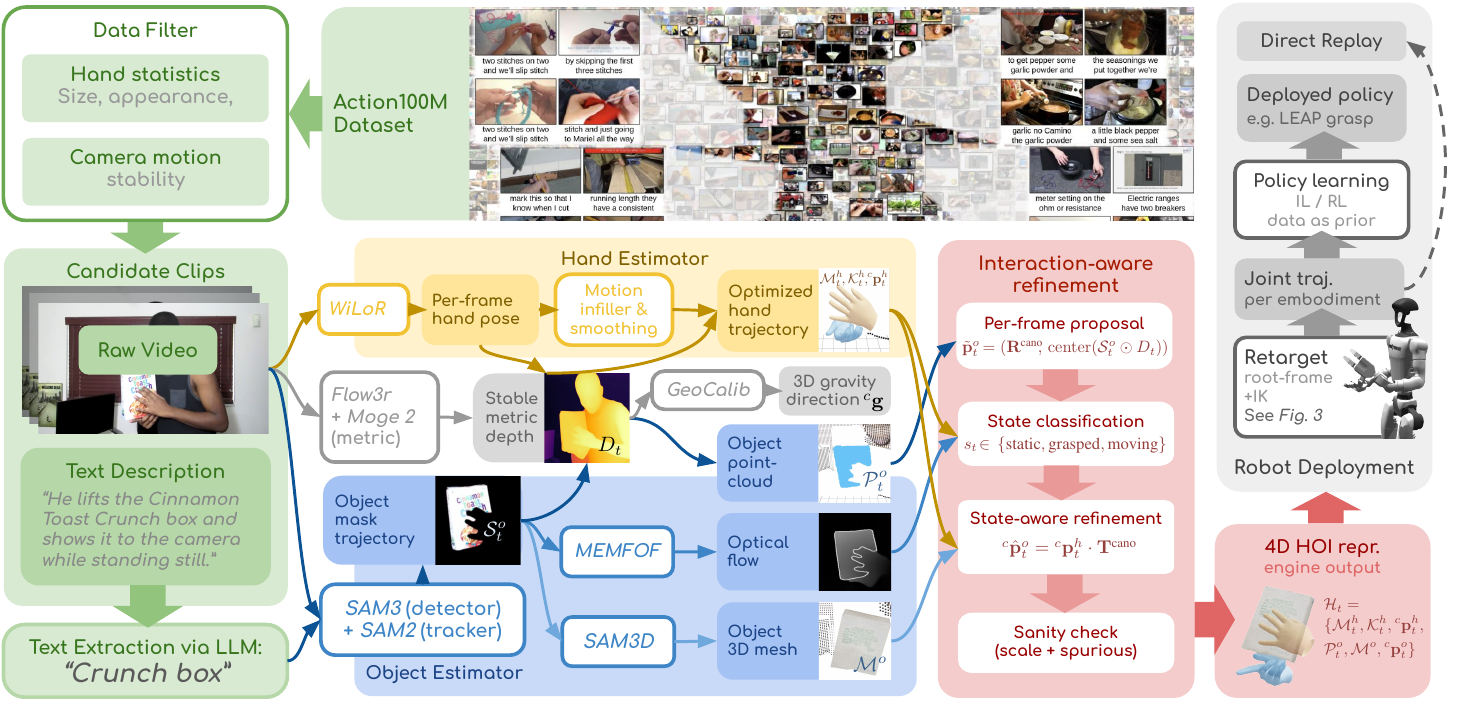}
  \captionof{figure}{\textbf{\egoinf pipeline.} From filtered in-the-wild Action100M clips and their text descriptions, the engine recovers metric hand trajectories and object geometry/pose for the automatically extracted objects. An interaction-aware refinement stage uses detected interaction states to align hand and object motion and suppress drift, yielding a metric, agent-agnostic 4D hand-object representation for downstream cross-embodiment retargeting and policy learning.}
  \label{fig:pipeline}
\end{center}

\begin{abstract}
Internet videos constitute the largest reservoir of embodied human manipulation knowledge, yet converting arbitrary RGB footage into actionable robot training data remains a major bottleneck. Existing lab- or factory-collected datasets are narrow in scale and diversity, limiting their ability to support open-world robot learning. Instead of proposing a static dataset, we introduce EgoInfinity, a universal 4D hand-object interaction data engine that enables web-scale data generation for robot retargeting and learning. EgoInfinity is built as a modular engine that integrates perception, segmentation, reconstruction, interaction-aware refinement, and retargeting components to automate this traditionally unscalable video-to-action problem without human-in-the-loop annotation. Its modular design allows the engine to continuously benefit from advances in any incorporated component. With EgoInfinity, in-the-wild human manipulation videos are lifted into agent-agnostic, metric 4D hand-object interaction representations, including hand trajectories, 6-DoF object poses, and contact-relevant states. Rather than naively connecting standalone components, EgoInfinity combines cross-module metric calibration with interaction-aware refinement to improve physical reliability, reducing drift and contact inconsistencies common in pure visual reconstruction. In addition, we propose a novel motion retargeter to compile the recovered 3D hand motions into executable joint trajectories for diverse robot morphologies, enabling video-to-action retargeting on any robot from arbitrary viewpoints and shot sizes (e.g., the human body is only partially visible). We validate EgoInfinity across perception fidelity, kinematic feasibility, contact consistency, cross-embodiment generalization, and real-robot skill acquisition (e.g., grasping, cutting, wiping, and pouring), demonstrating a scalable bridge from internet videos to executable robot behavior for robust open-world robot learning. Project Page: https://huggingface.co/spaces/Rice-RobotPI-Lab/EgoInfinity
\end{abstract}

% Two-to-three meaningful keywords (CoRL convention)
% \keywords{
%   Learning from Human Videos,
%   3D Hand-Object Interaction,
%   Cross-Embodiment Retargeting,
%   Automatic Data Generation
% }


% =====================================================================
% Body. 8-page limit applies HERE.
% =====================================================================
% =====================================================================
% sections/01_introduction.tex
%
% Compressed to ~1.25 pages for 8-page budget.
% Four paragraphs replace the previous seven; modularity demoted to a
% supporting design principle in §3.1 (per editorial decision Q2=B).
% No em-dashes, no bulleted lists. Contributions numbered inline.
% =====================================================================

\section{Introduction}
\label{sec:intro}

Training generalist robots requires manipulation data that is both scalable and diverse. 
Lab- or factory-collected data provides robot-compatible supervision, and egocentric demonstrations are especially valuable because they capture acting hands, manipulated objects, and contact patterns from an embodiment-aligned perspective. 
However, such data remains difficult to scale, and existing wearable or lab-collected corpora are limited by task diversity, collection setting, or participant behavior~\cite{hoque2025egodex, banerjee2025hot3d, zhan2024oakink2}. 
Internet videos offer a natural alternative, capturing humans manipulating everyday objects across diverse environments, viewpoints, and task contexts; \actiondataset alone contains 14.6 years of footage across 147M action segments~\cite{chen2026action100m}. 
Yet these videos are not directly robot-actionable: they lack metric 3D geometry, 6-DoF object state, contact information, and executable robot actions, causing existing learning-from-human-video methods to remain under-grounded for execution or mis-grounded when 2D action alignment is lifted into robot action space~\cite{ma2026robot, bharadhwaj2024track2act, wen2023any, bahl2023affordances, ye2025latent}. 
Therefore, automatically processing in-the-wild, web-scale videos into useful robot manipulation data, including curating egocentric-style data from arbitrary views, is a key step toward scalable robot learning.

We introduce \egoinf, a fully automated 4D manipulation data engine that converts in-the-wild RGB videos into agent-agnostic, metric hand-and-object representations without human annotation. 
It uses a modular pipeline spanning hand pose and mesh estimation, target-object discovery, object segmentation, monocular metric depth, camera/gravity estimation, object tracking and reconstruction, and interaction-aware refinement. 
This design makes \egoinf continuously upgradeable as individual modules improve. 
Importantly, \egoinf is not a naive composition of off-the-shelf components: it coordinates modules through unified coordinates, metric scale alignment, object-side priors, and interaction-aware refinement to reduce cross-module inconsistency and improve physical plausibility.

Furthermore, we provide a functional retargeting method that converts the processed 3D hand motions into executable robot motions across diverse embodiments. 
Rather than exactly replicating human body or arm kinematics, the retargeter estimates a robot-specific root transformation and preserves task-relevant hand motion within the target robot's constraints. 
This makes the approach applicable to arbitrary-view videos and partially observed humans, where the full body pose may be unavailable. 
Our contributions are: (i) a fully automated, modular 4D manipulation data engine for web-scale videos; (ii) cross-module calibration and interaction-aware refinement for more consistent hand-and-object reconstruction; and (iii) a functional cross-embodiment retargeter validated through kinematic feasibility, physical consistency, and real-robot skill acquisition.

\begin{table}[t]
  \centering
  \scriptsize
  \setlength{\tabcolsep}{5pt}
  \renewcommand{\arraystretch}{1.12}
  \resizebox{\linewidth}{!}{%
  \begin{tabular}{l l l c c c c}
    \toprule
    \textbf{Dataset} & \textbf{Source} & \textbf{Ann.} &
    \makecell{\textbf{Wear.}\\\textbf{req.?}} &
    \makecell{\textbf{Auto}\\\textbf{gen.?}} &
    \makecell{\textbf{Manual}\\\textbf{obj.?}} &
    \textbf{Scale} \\
    \midrule
    Ego4D~\cite{grauman2022ego4d}       & curated    & narr.    & headset & \ding{55} & \ding{55} & 3.7K hr \\
    EgoDex~\cite{hoque2025egodex}       & curated    & tracking & V. Pro  & \ding{55} & \ding{55} & 829 hr \\
    HOT3D~\cite{banerjee2025hot3d}      & curated    & mocap    & mocap   & \ding{55} & \ding{55} & 13.9 hr \\
    OakInk2~\cite{zhan2024oakink2}      & curated    & mocap    & mocap   & \ding{55} & \ding{55} & 6.5 hr \\
    UniHand-Mix~\cite{luo2026joint}     & aggregated & mixed    & partial & \ding{55} & \ding{55} & 1.2K--35K hr \\
    Open-X~\cite{o2024open}             & robot agg. & actions  & robot   & \ding{55} & \ding{55} & 1M+ traj. \\
    DROID~\cite{khazatsky2024droid}     & teleop     & actions  & robot   & \ding{55} & \ding{55} & 350 hr \\
    \midrule
    \textbf{\egoinf (ours)}
      & \textbf{internet}
      & \textbf{auto 4D}
      & \ding{51} none
      & \ding{51} full
      & \ding{51} none
      & \textbf{127K hr$^\dagger$} \\
    \bottomrule
  \end{tabular}
  }
  
  \vspace{2pt}
  \begin{minipage}{\linewidth}
    \raggedright
    \footnotesize$^\dagger$ Scale is currently bounded by \actiondataset; the data engine itself is corpus-agnostic.
  \end{minipage}

  \vspace{5pt}
  \caption{Comparison with prior egocentric manipulation and robot datasets.
  Columns indicate data source, annotation type, hardware requirement, automation level, manual object specification, and scale. \egoinf uniquely combines internet-scale data, no wearable requirement, full automatic generation, no manual object specification, and robot-usable 4D outputs.}
  \label{tab:data_quadrants}
\end{table}

\section{Related Work}
\label{sec:related}

\paragraph{Data Sources for Robot-Usable Manipulation.}
Existing manipulation data trades off scale, embodiment alignment, and robot actionability. 
As summarized in \tabref{tab:data_quadrants}, egocentric datasets such as Ego4D, EgoDex, HOT3D, and OakInk2 provide valuable human manipulation observations, but often rely on headsets, mocap, Vision Pro tracking, manual narration, or controlled collection~\cite{grauman2022ego4d, hoque2025egodex, banerjee2025hot3d, zhan2024oakink2}. 
DROID~\cite{khazatsky2024droid} and Open X-Embodiment~\cite{o2024open} provide executable robot actions but are limited by hardware, task design, and collection cost, while UniHand-Mix improves coverage through aggregation but inherits heterogeneous annotations and partial wearable dependence~\cite{luo2026joint}. 
In contrast, \egoinf targets in-the-wild internet videos and automatically extracts 4D manipulation data without wearables, depth sensors, CAD models, or human-specified object annotations, making it scalable like internet video while structured for robot use through metric hand motion, object state, and interaction cues.

\paragraph{From Human Videos to 4D Manipulation Data.}
Following the task/observation/action taxonomy of~\cite{ma2026robot}, task-oriented methods infer intent with VLMs~\cite{chen2024vlmimic, jain2024vid2robot, wake2024gpt}; observation-oriented methods bridge appearance gaps via video translation~\cite{lepert2025phantom, li2025h2r, lepert2025masquerade} or visual embeddings~\citep{nair2022r3m, ma2022vip, radosavovic2023real, cheang2024gr}; and action-oriented methods expose affordances or latent actions~\cite{chen2025villa, govind2026unilact}. 
\egoinf is closest to action-level, interaction-centric methods, but recovers metric hand-and-object structure rather than 2D pseudo-actions. 
This is newly feasible because key components have matured: RGB hand reconstruction~\cite{potamias2025wilor}, monocular metric geometry~\cite{wang2026moge, cong2026flow3r}, open-vocabulary segmentation and tracking~\cite{ravi2025sam, carion2025sam}, object reconstruction~\cite{sam3dteam2025sam3d3dfyimages}, optical-flow/PnP tracking~\cite{bargatin2025memfof}, and gravity/camera calibration~\cite{veicht2024geocalib}. 
% Unlike prior single-clip systems using subsets of these tools~\cite{zhang2025hawor, pavlakos2024reconstructing, wen2024foundationpose}, \egoinf integrates them with unified representations, cross-module calibration, automated target-object discovery, and interaction-aware refinement.
Recent single-clip systems make strong progress on world-grounded hand-object recovery from egocentric video~\cite{zhang2025hawor, pavlakos2024reconstructing, wen2024foundationpose, ye2026wholeworldgroundedhandobjectlifted}.
Most related to our perception design, the concurrent EgoGrasp~\cite{fu2026egograspworldspacehandobjectinteraction} reconstructs world-space hand-object interactions from dynamic egocentric video on a similar foundation-model stack and likewise uses grasp state to decide when hand motion drives object pose~\cite{shin2026composetrusthandsobject}, but recovers a world frame via SLAM-like camera estimation, relies on a full-body SMPL-X prior, and refines trajectories with learned diffusion priors.
\egoinf instead operates on approximately static-camera in-the-wild video without online SLAM or a body model, resolves grasp-state trust deterministically through interaction-state classification and rigid hand binding, and integrates these components with unified representations, cross-module calibration, and automated target-object discovery as a corpus-scale data engine.

\paragraph{Functional Cross-Embodiment Retargeting.}
Prior retargeting methods often target specific embodiment classes, such as dexterous hands~\cite{li2025maniptrans, chen2025vividex}, parallel grippers~\cite{kareer2025egomimic, park2025demodiffusion}, or humanoid upper bodies~\cite{qiu2025humanoid}. 
They work well when demonstrations and target embodiments are aligned, but arbitrary internet videos often contain only hands, partial arms, or changing viewpoints, making exact body-pose recovery and kinematic imitation unreliable. 
\egoinf instead exposes an agent-agnostic 4D manipulation representation and performs functional retargeting: it estimates a feasible robot-specific root transformation and preserves task-relevant hand motion within the target robot's constraints, enabling the same recovered video data to be compiled into executable trajectories for diverse morphologies.

\section{The \egoinf Data Engine}
\label{sec:engine}

% \begin{figure}[t]
%   \centering
%   \includegraphics[width=\columnwidth]{figures/pipeline.pdf}
%   \caption{\textbf{\egoinf pipeline.} From filtered in-the-wild Action100M clips and their text descriptions, the engine recovers metric hand trajectories and object geometry/pose for the automatically extracted objects. An interaction-aware refinement stage uses detected interaction states to align hand and object motion and suppress drift, yielding a metric, agent-agnostic 4D hand-object representation for downstream cross-embodiment retargeting and policy learning.}
%   \label{fig:pipeline}
%   \vspace{-10pt}
% \end{figure}

\egoinf processes raw internet RGB videos and their semantic annotations into structured 4D manipulation data. 
The engine outputs metric hand states, object point clouds and meshes, 6-DoF object pose trajectories, contact-relevant states, and coordinate-reframed outputs for downstream retargeting and learning. 
A schematic overview is shown in \figref{fig:pipeline}.

\subsection{Modular Pipeline Architecture}
\label{sec:engine:arch}

% \egoinf is designed as a modular pipeline for converting internet RGB videos into structured 4D manipulation data. 
% It integrates hand mesh estimation, metric geometry recovery, camera and gravity calibration, target-object discovery, object reconstruction and tracking, and interaction-aware refinement within a unified intermediate representation. 
% This modular design makes the engine component-replaceable: each module can be upgraded independently as stronger foundation or geometry models become available. 
% For scalable processing, \egoinf uses a two-pass strategy: \emph{Pass 1} performs a lightweight temporal scan to identify hand-present segments, while \emph{Pass 2} runs the full reconstruction stack only on these active segments. 
% We currently target internet videos with approximately static views, which are common in tutorial and how-to manipulation content.

\egoinf is designed as a modular pipeline for converting internet RGB videos into structured 4D manipulation data. 
It integrates hand mesh estimation, metric geometry recovery, camera and gravity calibration, target-object discovery, object reconstruction and tracking, and interaction-aware refinement within a unified intermediate representation. 
This modular design makes the engine component-replaceable: each module can be upgraded independently as stronger foundation or geometry models emerge. 
For scalability, \egoinf uses a two-pass strategy: \emph{Pass 1} performs a lightweight temporal scan to identify hand-present segments and filter videos using hand-motion statistics and camera-motion cues, retaining clips likely to contain useful manipulation. 
\emph{Pass 2} runs the full reconstruction stack only on active segments. 
We currently target internet videos with approximately static views, common in tutorial and how-to manipulation content.

\subsection{Metric-Calibrated Hand--Object Tracking}
\label{sec:engine:tracking}

\textbf{Unified Metric Geometry.}
A central design principle of \egoinf is cross-module calibration: rather than consuming each model's raw output, we align modules along a shared metric scale, a unified camera frame, object-side geometric priors, and interaction-aware refinement. Hand and object predictions are produced by different modules and are not inherently aligned in scale or 3D reference frame. To enable reliable hand--object association, contact reasoning, and downstream retargeting, \egoinf first calibrates them into a shared metric camera-frame geometry from RGB video. Specifically, we use \moge~\cite{wang2026moge} to estimate camera focal length and global metric scale, \flowthreer~\cite{cong2026flow3r} to predict dense depth maps, and \geocalib~\cite{veicht2024geocalib} to estimate the gravity vector ${}^{c}\gravity$. This shared metric reconstruction lifts image-space hand and object predictions into the same calibrated 3D space, providing the geometric basis for consistent hand poses, object point clouds, and downstream pose tracking.

\textbf{Metric 3D Hand Tracking.}
We use \wilor~\cite{potamias2025wilor} to estimate MANO hand parameters~\cite{MANO:SIGGRAPHASIA:2017}, including pose $\thand_t$ and shape $\bhand_t$, from RGB frames. These parameters define hand meshes $\{\mathcal{M}^h_t\}_{t=1}^{T}$, 3D hand keypoints $\{\mathcal{K}^h_t\}_{t=1}^{T}$, and global hand poses ${}^{c}\pose^h_t = ({}^{c}\mathbf{R}^h_t, {}^{c}\mathbf{t}^h_t) \in SE(3)$. An infilling module completes missing or unstable hand estimates, and the recovered hand trajectory is grounded in the shared metric geometry above. The resulting camera-frame hand motion is therefore calibrated with object geometry under the same scale, supporting reliable object association.
% , physical refinement, and robot retargeting.

\textbf{Object Discovery, Reconstruction, and Tracking.}
The manipulated object is discovered automatically from semantic prompts and visual evidence, without human annotation. We use the video description as a semantic prompt for \samthree~\cite{carion2025sam} detection, then propagate the detected mask through the segment using \samtwo~\cite{ravi2025sam} and lift it with depth to produce per-frame object point clouds $\{\mathcal{P}^o_t\}_{t=1}^{T}$. When sufficient visual evidence is available, \samthreed~\cite{sam3dteam2025sam3d3dfyimages} reconstructs the object mesh $\mathcal{M}^o$. Given the object mesh, masks, and point clouds, \foundationposepp~\cite{Wenhao_Yan_and_Jie_Chu_FoundationPose_2025} tracks the object 6-DoF pose trajectory $\{{}^{c}\pose^o_t\}_{t=1}^{T}$ in the same metric frame as the hands, with additional stabilization for static or weakly observed frames. Together, these components produce the initial 4D manipulation state
\[
    \mathcal{H}_t =
    \{\mathcal{M}^h_t, \mathcal{K}^h_t, {}^{c}\pose^h_t,
    \mathcal{P}^o_t, \mathcal{M}^o, {}^{c}\pose^o_t\}.
\]

% \subsection{\phaseF: Physics-Aware Refinement}
% \label{sec:engine:phaseF}

% Pure visual reconstruction can be visually plausible but physically unreliable: hand meshes may be misaligned with object geometry, object poses may drift, and contact intervals may be inconsistent. 
% Given the initial sequence $\mathcal{H}_{1:T}$, \phaseF refines it into a more physically grounded trajectory,
% \[
%     \hat{\mathcal{H}}_{1:T}
%     =
%     \mathcal{F}_{\mathrm{phys}}(\mathcal{H}_{1:T}).
% \]
% The refinement combines contact detection, geometric penetration correction, object-pose refinement, static locking for non-grasped objects, and temporal smoothing. 
% For object motion, FoundationPose++ provides neural 6-DoF refinement, FlowPnP/PnP uses 2D--3D correspondences when reliable, PCA/OBB anchors provide object-side orientation and scale priors, and \texttt{fp\_compose} composes frame-to-frame updates into a temporally consistent object trajectory. 
% These steps reduce drift and improve physical plausibility before downstream retargeting and learning.

\subsection{Interaction-Aware Refinement}
\label{sec:engine:refinement}

Pure visual object tracking can be temporally unstable: poses may drift for static objects, fail under hand occlusion, or fluctuate when visual correspondences are weak. Rather than relying on visual tracking alone, we use detected interaction states to refine object trajectories. Each frame first receives an initial 6-DoF proposal $\tilde{\pose}^o_t=(\mathbf{R}^{\text{cano}},\,\operatorname{center}(\mathcal{S}^o_t\odot D_t))$ from the object mask and depth, keeping the canonical \samthreed orientation $\mathbf{R}^{\text{cano}}$ and estimating translation from the back-projected masked points. We then classify each frame from \memfof~\cite{bargatin2025memfof} optical flow and hand keypoints as $s_t\in\{\text{static},\text{grasped},\text{moving}\}$. Based on $s_t$, a grasped object is rigidly attached to the hand frame as ${}^{c}\hat{\pose}^o_t={}^{c}\pose^h_t\cdot\mathbf{T}^{\text{cano}}$ with a palm-aligned canonical transform $\mathbf{T}^{\text{cano}}$; a static object is locked to its robust point-cloud centroid; and a moving object retains the proposal $\tilde{\pose}^o_t$. Finally, we apply lightweight sanity checks to suppress implausible scales and spurious background detections; these three stages correspond to the refinement block in \figref{fig:pipeline}. We expose finer-grained interaction states in our implementation; additional details are provided in Appendix~A.

\subsection{Output Cleanup and Coordinate Reframing}
\label{sec:engine:output}

For consumers that require clean object point clouds, \egoinf optionally applies mask erosion, depth-gradient filtering, and statistical outlier removal. 
Finally, under the approximately static-camera setting, exo-to-ego conversion is performed as a rigid coordinate reframing in recovered 3D space rather than as 2D generative video translation. 
This avoids pixel hallucination from inpainting or GAN-based translation~\cite{lepert2025phantom, li2025mimicdreamer, ci2025h2r} while preserving geometric consistency.

\section{Cross-Embodiment Robot Motion Retargeting}
\label{sec:retargeting}

Retargeting converts agent-agnostic 3D hand motion into robot-specific motion. 
Rather than requiring an exact human body pose or full kinematic motion, we estimate a feasible root transformation for the target embodiment, enabling transfer across robot morphologies. 
This is crucial for internet videos where only the hands are visible and the arms or body are partially observed or absent. 
Our goal is functional retargeting: preserving task-relevant hand motion without mimicking precise human body or arm motion.

\subsection{Equivariant Neural Estimation of the Kinematic Root Frame}

Given reconstructed hand trajectories $\{({}^{c}\Rmat^h_t, {}^{c}\tvec^h_t)\in SE(3)\}_{t=1}^T$ and optional gravity ${}^{c}\gravity$, we use a neural network $\Phi(\cdot)$ to estimate a shared kinematic root frame, e.g., a humanoid torso frame. 
% The network outputs ${}^{c}\pose^r=({}^{c}\mathbf{R}^r,{}^{c}\mathbf{t}^r)\in SE(3)$ in the camera frame, transforming recovered hand motions into root-relative coordinates for retargeting.
The network predicts ${}^{c}\pose^r = ({}^{c}\mathbf{R}^r, {}^{c}\mathbf{t}^r) \in SE(3)$ in the camera frame, which converts the recovered hand motion into root-relative coordinates for retargeting, as illustrated in \figref{fig:retarget}. More architectural details of this network $\Phi$ are given in Appendix~B.

\textbf{Vector-Neuron $SE(3)$-Equivariant Architecture.}
We design $\Phi$ to be $SE(3)$-equivariant: if the input trajectories are rigidly transformed, or the camera frame changes, the predicted root frame transforms accordingly. 
For any $\bm{G}\in SE(3)$ applied to observations $\mathbf{x}$,
\begin{equation}
    \bm{G}\cdot\Phi(\mathbf{x}) = \Phi(\bm{G}\cdot\mathbf{x}),
\end{equation}
where $\mathbf{x}$ contains the hand trajectories and optional gravity. 
We implement this mapping with Vector Neuron (VN) layers~\cite{deng2021vector}, which process 3D vector features and are rotation-equivariant by construction. 
Translation equivariance is obtained by centering hand positions around their centroid $\bm{c}$. 
Each timestep is encoded by centered hand positions, hand orientation axes, and optional gravity; bilateral trajectories are fused with VN linear layers, processed by a VN-Transformer encoder, and temporally pooled.

The rotation head decodes two vector outputs into ${}^{c}\mathbf{R}^r\in SO(3)$ via Gram--Schmidt orthogonalization. 
For translation, the network predicts a root-relative offset $\mathbf{v}$ and maps it back as ${}^{c}\mathbf{t}^r = {}^{c}\mathbf{R}^r\mathbf{v}+\mathbf{c}$, avoiding direct camera-frame translation regression while preserving equivariance.

\textbf{Flow-Matching Estimation.}
Instead of deterministic regression, we formulate root-frame prediction as flow-matching conditional generation~\cite{lipman2022flow}, modeling $p({}^{c}\pose^r\mid\mathbf{x})$ over plausible root frames. 
This captures cases where different torso poses yield the same hand motion, especially under partial-body observations. 
From prior samples $({}^{c}\mathbf{R}^r_0,{}^{c}\mathbf{t}^r_0)$, with ${}^{c}\mathbf{R}^r_0\sim\mathcal{U}(SO(3))$ and ${}^{c}\mathbf{t}^r_0\sim\mathcal{N}(\mathbf{c},0.5^2\mathbb{I})$, the learned flow maps to root-frame hypotheses conditioned on $\mathbf{x}$. 
At inference, samples are generated by integrating the learned ODE with 20 Euler steps.

\begin{figure}[t]
  \centering
  \includegraphics[width=\columnwidth]{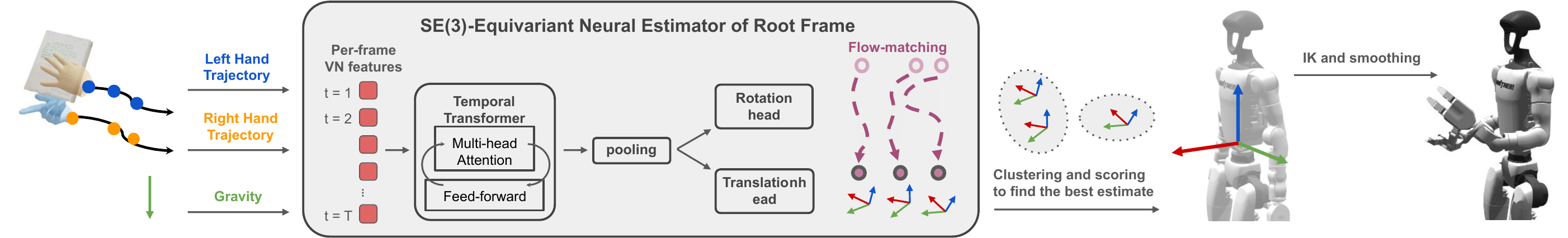}
  % \caption{\textbf{\egoinf pipeline.} human video --> text prompt --> hand and object 4D output --> retargeter --> estimate torso frame --> motions.}
  \caption{Retargeting pipeline. Recovered 3D hand trajectories and gravity are fed into a simulation-trained, robot-specific root estimator $\Phi$, which generates candidate root frames. After clustering and scoring, the best candidate (e.g., the torso frame for Unitree G1) is selected, and the hand motion is retargeted through IK followed by post-optimization.}
  \label{fig:retarget}
  \vspace{-10pt}
\end{figure}

\subsection{Robot-Specific Model Training}

The network is trained entirely in MuJoCo simulation~\cite{todorov2012mujoco}, without real-world supervision. 
For each target robot, we procedurally generate paired hand trajectories and ground-truth root poses. 
Per-arm Cartesian hand trajectories are sampled around the manipulation workspace: starting from a noisy reference joint configuration, forward kinematics provides a hand anchor; smooth control points are generated by an Ornstein--Uhlenbeck random walk biased toward this anchor, converted to joint-space knots with warm-started position-only IK, and interpolated with a cubic spline. 
For each trajectory, a random camera pose is sampled around the robot and used to transform both hand trajectories and root frames into the camera frame.

\textbf{Augmentations.}
To improve robustness to in-the-wild reconstruction errors, we apply tracking noise, tracking jumps, hand occlusion, and gravity noise. 
These augmentations mimic noisy hand-pose estimates, temporary tracker failures, partial or single-hand observations, and inaccurate or missing gravity estimates. 
We additionally drop gravity for $30\%$ of samples so the network degrades gracefully when gravity is unavailable.

\subsection{Trajectory Post-Processing and Robot Retargeting}

% \textbf{Root-Trajectory Hypothesis Selection.}
% At inference, we divide the video into overlapping 2-second windows and sample root-frame hypotheses $\{({}^{c}\mathbf{R}^r_k,{}^{c}\mathbf{t}^r_k)\}$ from the flow model. 
% Hypotheses are clustered in rotation space using geodesic $k$-means, and $K=5$ representatives are retained. 
% Since the camera or human body may move, window-level estimates are treated as keyframes and interpolated into smooth per-frame root trajectories using linear interpolation for translation and SLERP for rotation. 
% Each candidate is evaluated through downstream IK: hand poses are transformed into the time-varying root frame and tracked frame-by-frame with warm starts and null-space regularization. 
% Candidates are scored by IK convergence, residual tracking error, manipulability, joint-limit margin, and smoothness; the best one is selected.

\textbf{Root-Trajectory Hypothesis Selection.}
At inference, we divide the video into overlapping temporal windows and sample multiple root-frame hypotheses $\{({}^{c}\mathbf{R}^r_k,{}^{c}\mathbf{t}^r_k)\}$ from the flow model. Hypotheses are clustered by $k$-means, and a small set of representative candidates is retained. Since either the camera or the human body may move, these window-level estimates are treated as keyframes and interpolated into smooth per-frame root trajectories using linear interpolation for translation and SLERP for rotation. Each candidate is then evaluated through downstream IK: hand poses are transformed into the time-varying root frame and tracked frame-by-frame with warm starts and null-space regularization. Candidates are scored by IK convergence, residual tracking error, manipulability, joint-limit margin, and smoothness, and the best one is selected. More implementation details are provided in Appendix~B.

\textbf{Motion and Hand Post-Processing.}
The selected joint trajectory is refined by interpolating failed IK frames and applying repeated uniform filtering to remove high-frequency jitter. 
For dexterous hands, finger joints are retargeted separately from reconstructed MANO hand keypoints using a geometry-based, robot-specific mapping. 
Thus, arm joints follow wrist-level IK targets, while finger joints are inferred directly from hand keypoints.
% \input{sections/05_dataset}
% =====================================================================
% sections/06_experiments.tex
% Compressed from 6 to 4 subsections (target: 1.0 page).
% Each subsection still states explicitly *why* the measured property
% predicts real-robot success, satisfying the CoRL transferability
% criterion.
% No em-dashes, no bulleted lists. Configurations described in prose.
% =====================================================================

\section{Experiments and Validation}
\label{sec:experiments}

\begin{figure}[t]
  \centering
  \includegraphics[width=\columnwidth]{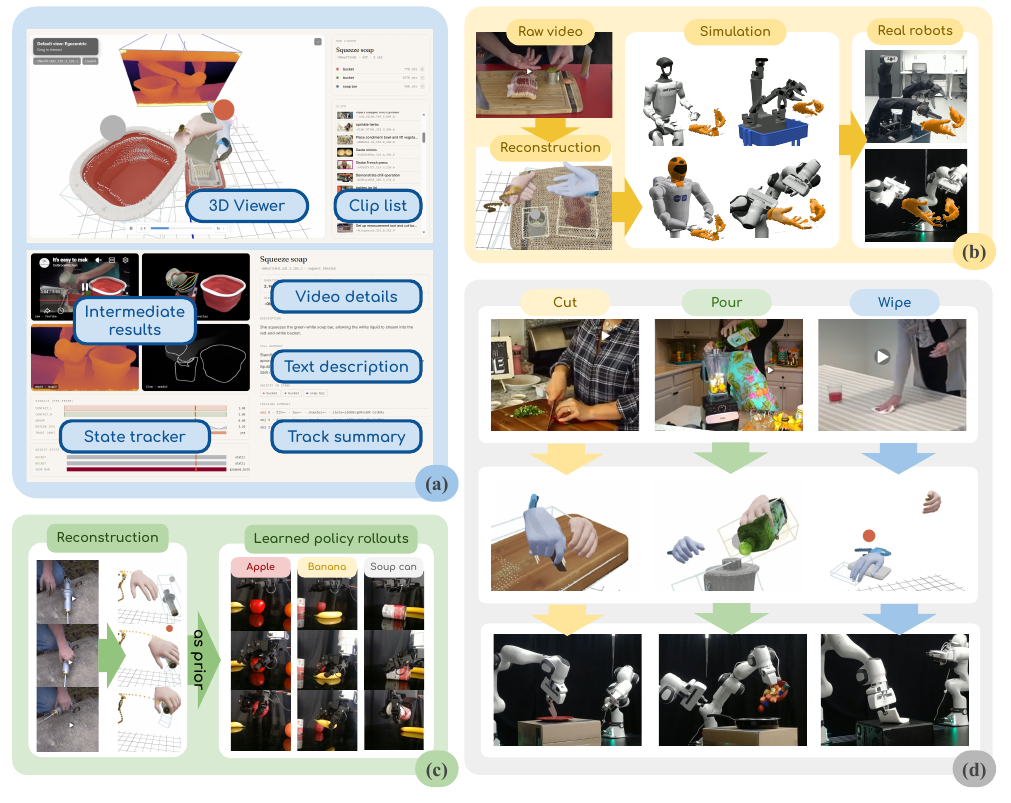}
  \vspace{-10pt}
  \caption{\textbf{\egoinf experiments.} (a) Project page visualization (3D viewer, intermediate results, text descriptions, track summaries). (b) 4D HOI reconstructions retargeted to multiple embodiments in simulation and on real robots. (c) Extracted hand trajectories used as priors for downstream policy use, generalizing across objects. (d) Real-robot demos on Cut, Pour, and Wipe.}
  \label{fig:experiment}
    \vspace{-5pt}
\end{figure}

We evaluate \egoinf from four perspectives: interactive data access, curated dataset statistics, cross-embodiment retargeting, and real-robot execution and learning.

\subsection{Browser-Based Data Server and Visualization}
\label{sec:experiments:server}

We develop an online browser-based data server for searching, processing, visualizing, and downloading \egoinf-generated 4D manipulation data. 
As shown in \figref{fig:experiment} (a), users can search the internet-video corpus with natural-language keywords, inspect retrieved clips, launch online processing, and visualize intermediate and final outputs, including MANO hand mesh trajectories, 6-DoF object trajectories, reconstructed object geometry, point clouds, 3D bounding boxes, depth maps, optical flow, and detected grasp events. 
The interface makes outputs directly inspectable for failure diagnosis and turns \egoinf into an interactive curation tool.
% for downstream robot learning and retargeting.
\vspace{-10pt}

\subsection{Curated \actiondataset Dataset}
\label{sec:experiments:dataset}

\vspace{-10pt}

% We curate an extensible subset of Action100M with 100 processed manipulation videos hosted on our data server. Each video is processed by \egoinf to produce downloadable 4D outputs.
% % , including metric hand trajectories, object masks, object point clouds, reconstructed object geometry when available, 6-DoF object trajectories, grasp/contact events, depth maps, and optical flow. 
% We report dataset statistics in \figref{fig:dataset_stats}, including task and object categories, video duration, hand-present frames, detected interaction segments, object-track coverage, reconstruction success rate, and grasp-event frequency.

\begin{figure}[t]
  \centering
  \includegraphics[width=\columnwidth,height=0.18\textheight,keepaspectratio]{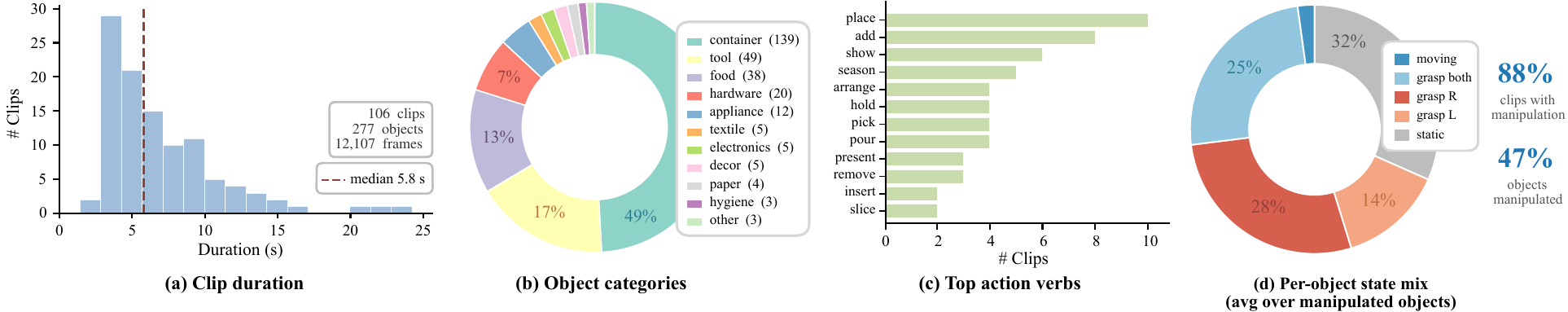}
  \caption{\textbf{Statistics of the curated \actiondataset subset.} (a) Clip durations. (b) Object category mix. (c) Top action verbs. (d) Per-frame state distribution averaged across manipulated objects (d). 88\% of clips and 47\% of objects are manipulated, with balanced use of left, right, and bimanual grasps.}
  \label{fig:dataset_stats}
  \vspace{-20pt}
\end{figure}

We curate an extensible subset of \actiondataset with 106 processed manipulation videos hosted on our data server. 
Each video is processed by \egoinf to produce downloadable 4D outputs. 
We report dataset statistics in \figref{fig:dataset_stats}, including clip duration, object categories, top action verbs, and per-frame interaction-state distribution across manipulated objects. 
More examples from the curated dataset are provided in Appendix~C.

\subsection{Cross-Embodiment Motion Retargeting}
\label{sec:experiments:retargeting}

% We evaluate retargeting across three substantially different embodiments: Unitree G1, NASA's Robonaut2, and a dual-Franka FR3 setup. For each robot, the retargeter estimates a robot-specific root transformation and solves IK to convert recovered 3D hand motions into executable joint trajectories under the robot's kinematic constraints. We report dataset-level statistics in \tabref{tab:retargeting_stats}, including IK success rate, residual hand-pose error, joint-limit margin, manipulability, and trajectory smoothness, using retargeting feasibility as both a system evaluation and a metric for how well each robot design supports human hand motions.

We evaluate retargeting on three substantially different embodiments: Unitree G1, NASA's Robonaut2, and a dual-Franka FR3 setup. 
For each robot, the retargeter estimates a robot-specific root transformation and solves IK to convert recovered 3D hand motions into executable joint trajectories under the robot's kinematic constraints. 
We report dataset-level statistics in \tabref{tab:retargeting_stats}, including IK success rate, residual hand-pose tracking error, joint-limit margin, manipulability, and trajectory smoothness. 
\figref{fig:experiment} (b) shows an example motion retargeted across multiple embodiments in simulation and deployed on real robots, with additional retargeted examples provided in Appendix~C.

\begin{table}[t]
\centering
% \scriptsize
% \scriptsize
\setlength{\tabcolsep}{4pt}
\begin{tabular}{lcccccc}
\toprule
Robot
  & IK Rate
  & Pos.\ Error
  & Ori.\ Error
  & Jnt.-Limit Margin
  & Manipulability
  & Smoothness \\
\midrule
Unitree G1
  & $0.821$
  & $2.86$ cm
  & $6.73$°
  & $0.619$ rad
  & $0.012$
  & $0.00693$ \\
Robonaut2
  & $0.774$
  & $6.67$ cm
  & $8.25$°
  & $0.134$ rad
  & $0.058$
  & $0.00343$ \\
Dual-Franka
  & $0.706$
  & $10.27$ cm
  & $12.17$°
  & $0.572$ rad
  & $0.080$
  & $0.00582$ \\
\bottomrule
\end{tabular}
\vspace{5pt}
\caption{
  IK Rate: per-frame IK success rate.
  Pos./Ori.\ Error: mean hand position ($\ell_2$, cm) and orientation (geodesic, °) error between IK target and achieved pose.
  Jnt.-Limit Margin: mean minimum joint clearance (rad).
  Manipulability: mean manipulability index $\sqrt{\det(JJ^\top)}$.
  Smoothness: mean squared joint velocity $\dot q$.
}
\label{tab:retargeting_stats}
\vspace{-15pt}
\end{table}

\vspace{-5pt}

\subsection{Real-Robot Retargeting and Skill Learning}
\label{sec:experiments:real_robot}

\vspace{-5pt}

% We further test whether \egoinf-generated motions support real-robot learning and execution. First, we use extracted hand motions as priors to train a grasping policy on a real LEAP dexterous hand, enabling grasping of diverse objects. Second, we directly retarget extracted motions to a real Franka FR3 robot, showing functional execution of swiping, pouring, and cutting skills. These results evaluate the full video-to-action pipeline, from in-the-wild RGB videos to 4D manipulation data, robot-specific retargeting, policy learning, and real-world execution.

We further test whether \egoinf-generated motions support real-robot learning and execution. 
First, we use extracted hand motions as priors to train a grasping policy on a real LEAP dexterous hand, enabling grasping of diverse objects (\figref{fig:experiment} (c)). 
Second, we directly retarget extracted motions to a real dual-arm Franka FR3 setup, demonstrating functional execution of cutting, pouring, and wiping skills (\figref{fig:experiment} (d)). 
These results validate the full video-to-action pipeline, from in-the-wild RGB videos to 4D manipulation data, robot-specific retargeting, downstream policy use, and real-world execution. 
Additional real-robot examples are provided in Appendix~C.

\section{Conclusion}
\label{sec:conclusion}
\vspace{-5pt}

We presented \egoinf, a universal 4D manipulation data engine that converts pure-RGB internet videos into agent-agnostic, robot-usable hand-and-object representations without mocap, depth sensors, CAD models, wearables, or human-specified object annotations. By prioritizing metric 3D geometry, 6-DoF object state, and interaction-aware refinement over 2D pseudo-actions, \egoinf improves the physical grounding of data extracted from human videos. By leveraging unscripted, equipment-free internet videos, it also captures diverse manipulation behaviors, pedagogical demonstrations, and natural multimodal context that are difficult to reproduce in curated lab datasets. Together with functional cross-embodiment retargeting, \egoinf provides scalable infrastructure for turning web-scale human video into executable robot behavior, supporting future work in robot learning, multimodal grounding, and HRI. 

% =====================================================================
% sections/07_limitations.tex
%
% NEW in v3, required by CoRL 2026:
%   "All submissions should include a Limitations section (counted
%    toward the 8-page limit), explicitly describing limiting
%    assumptions, failure modes, and other limitations of the results
%    and experiments, and how these might be addressed in the future."
%
% Target: 0.5 page. Pure prose; no em-dashes; no bulleted lists.
% =====================================================================

\section{Limitations}
\label{sec:limitations}
\vspace{-5pt}

\egoinf currently assumes approximately static-camera videos, excluding body-mounted or hand-held footage. 
This makes web-scale processing tractable and avoids online SLAM, but limits corpus diversity; as SLAM-aware depth models mature, our modular architecture can incorporate stronger geometry modules to relax this assumption. 
\egoinf also does not solve precise physical hand-object alignment. 
Its interaction-aware refinement provides coarse grasp detection and spatially correlates hand and object trajectories, but does not guarantee contact-level accuracy such as exact fingertip placement, force consistency, or no-slip constraints. 
In addition, \egoinf does not provide tactile observations, which are important for fine-grained contact reasoning and have been explored in tactile-specific datasets~\citep{zhou2026touchanything}. 
Finally, the motion retargeter is robot-specific and may require retraining or calibration for new robot designs. 
It targets functional motion transfer rather than fine-grained kinematic imitation, and may be insufficient for tasks requiring exact hand posture, precise contact timing, tactile feedback, or highly dexterous manipulation.

% =====================================================================
% Acknowledgments. Auto-rendered only in final / preprint modes;
% silently dropped in the anonymized initial submission.
% =====================================================================
% \acknowledgments{%
%   Acknowledgments will be added in camera-ready.%
% }

% =====================================================================
% Bibliography. The corl style sets bibliographystyle automatically.
% =====================================================================
\bibliography{references}

% =====================================================================
% Appendix. Does NOT count toward the 8-page main-paper limit.
% =====================================================================
\clearpage
\appendix
\section*{Appendix: Implementation and Data Details}
This supplementary material is organized into three parts. \secref{app:engine}
details the data engine of \secref{sec:engine}, including the egocentric view
synthesis that gives \egoinf its name, the full interaction-state definition,
the rigid hand-binding mechanism, the robust-statistics primitives, the
off-the-shelf perception-module configuration, our coordinate conventions, the sanity
filtering of implausible detections, and a consolidated
table of all (non-learned) thresholds. \secref{app:retarget} covers the
cross-embodiment retargeter of \secref{sec:retargeting}, including its network
architecture, training setup, and sliding-window inference. \secref{app:results}
provides additional qualitative results, from the curated egocentric
reconstructions and the interactive browser through the full video-to-robot
pipeline to the downstream grasping policy. 

\section{Data Engine Implementation Details}
\label{app:engine}

This appendix expands the data engine of \secref{sec:engine}. It details the
egocentric view synthesis behind \egoinf's name, the full interaction-state
definition underlying the three-way description of \secref{sec:engine:refinement},
the state-dependent pose sources, the rigid hand-binding mechanism, the
robust-statistics primitives reused throughout, the configuration of the
off-the-shelf perception modules, our coordinate conventions, and a consolidated
table of all (non-learned) thresholds.

\begin{figure}[h]
  \centering
  \includegraphics[width=\columnwidth]{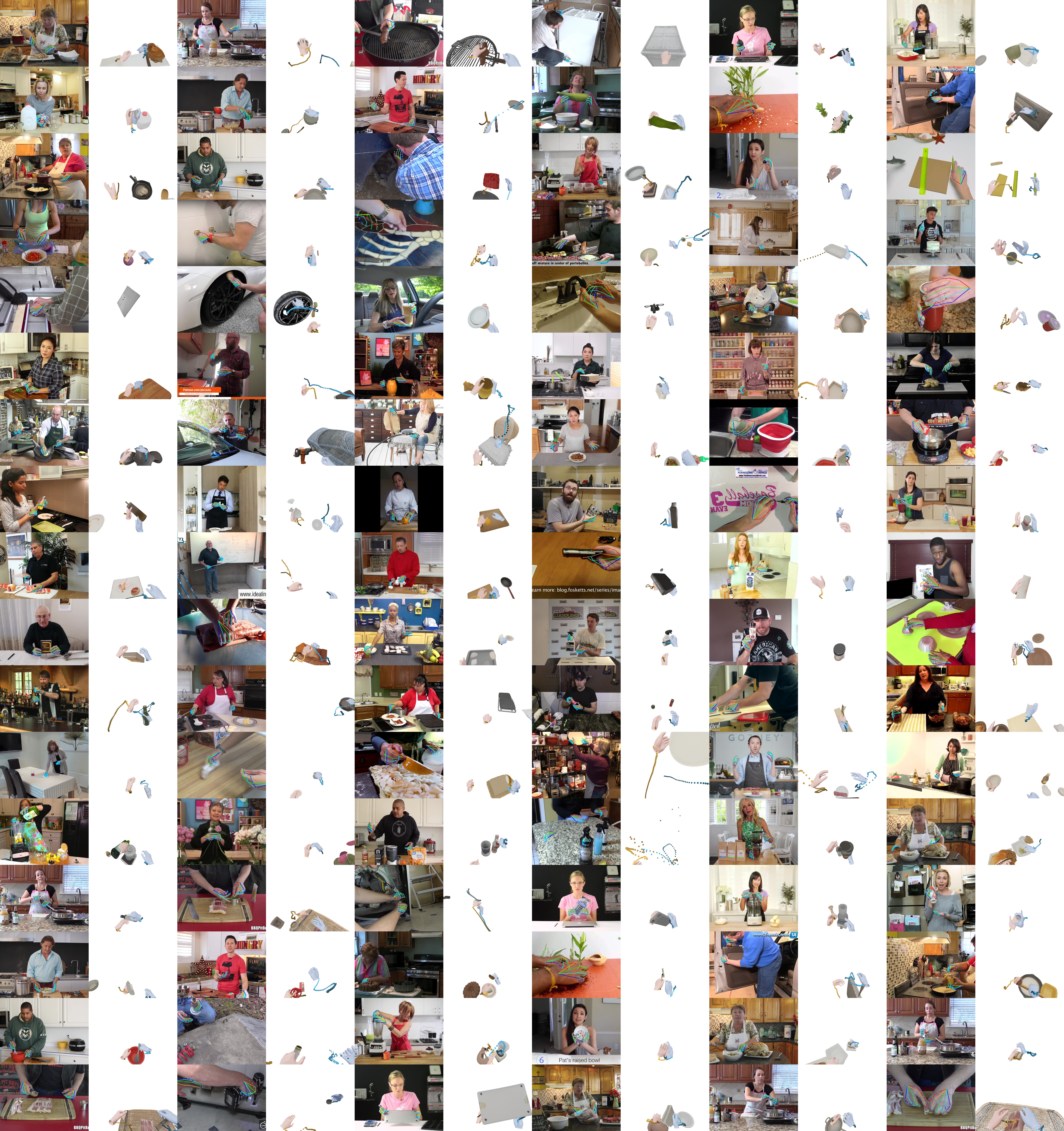}
  \caption{Curated \actiondataset reconstructions in synthesized
  egocentric view. Each pair shows the original exocentric source frame
  (left) and the corresponding egocentric re-rendering (right) with
  recovered hand meshes, object meshes and point clouds, and 6-DoF poses,
  produced by the gravity-aligned interaction-following camera described
  above. The source clips share no common capture viewpoint, yet all are
  reframed into a consistent egocentric observation.}
  \label{fig:app_dataset}
\end{figure}
% ---------------------------------------------------------------------
\subsection{Egocentric View Synthesis}
\label{app:engine:ego}
% ---------------------------------------------------------------------
A defining capability of \egoinf is turning arbitrary-viewpoint source
video into a consistent egocentric observation, which is what its name
refers to. The exo-to-ego conversion of \secref{sec:engine:output} is a
rigid reframing of the recovered 3D scene, not a 2D generative
translation. Since all hand and object geometry already lives in the
metric camera-world frame of \secref{app:engine:coords}, an egocentric
view is obtained by re-rendering from a synthesized ego camera: for each
frame we place a virtual camera at a fixed offset above a hand-derived
anchor (the bilateral hand midpoint), set its up-axis to the GeoCalib
gravity vector ${}^{c}\gravity$, and aim it at the hand-object
interaction region, following the anchor frame by frame. The reframing
thus reduces to applying the ego extrinsics to the reconstructed points
${}^{c}\mathbf{p}_t$, preserving metric geometry and contact exactly.

This interaction-centric placement is deliberate rather than an attempt
to reproduce a true head view. Against the raw exocentric frames, it
keeps the interaction centered, close, and viewpoint-consistent across
clips; against a hypothetical recovered head trajectory, it avoids the
frames a real head wastes by turning away or being occluded, retaining
them as usable egocentric observations and so raising the usable-frame
yield per clip. We emphasize this is a deterministic, functional
approximation that does not recover the true head pose (consistent with
the partial-observation setting of \secref{sec:limitations}), mirroring
the \emph{functional over exact} principle of our retargeter
(\secref{sec:retargeting}). This embodiment-aligned unification of
arbitrary viewpoints into a consistent egocentric observation, at corpus
scale, is the sense in which \egoinf operates. Example reconstructions in
this synthesized egocentric view are shown in \figref{fig:app_dataset}
(further discussed in \secref{app:results:dataset}).

\subsection{Full Interaction-State Definition}
\label{app:engine:states}
% ---------------------------------------------------------------------
\secref{sec:engine:refinement} classifies each object frame into the
three states $s_t \in \{\textsc{static}, \textsc{grasped}, \textsc{moving}\}$
that are sufficient to describe the refinement logic. Internally, the
engine uses a finer six-state label that distinguishes scene-fixed
objects from momentarily resting ones and resolves which hand is in
contact:
\begin{equation}
  \sigma_t \in \Sigma = \{
    \textsc{static\_global},\;
    \textsc{static},\;
    \textsc{grasped\_l},\;
    \textsc{grasped\_r},\;
    \textsc{grasped\_both},\;
    \textsc{moving}
  \}.
\end{equation}
The mapping to the main-text states is
$\{\textsc{static\_global}, \textsc{static}\}\!\to\!\textsc{static}$,
$\{\textsc{grasped\_l}, \textsc{grasped\_r}, \textsc{grasped\_both}\}\!\to\!\textsc{grasped}$,
and $\textsc{moving}\!\to\!\textsc{moving}$.

\paragraph{Global static gate.}
Let $\mathbf{c}_t \in \mathbb{R}^2$ be the object mask centroid in image
coordinates and $\Delta_{[10,90]} = \lVert \mathrm{p}_{90}(\mathbf{c}) - \mathrm{p}_{10}(\mathbf{c}) \rVert_2$
its inter-percentile span over the clip. If
$\Delta_{[10,90]} \le 0.02 \cdot \min(H,W)$ the object is declared
\textsc{static\_global} for the entire clip. This short-circuit handles
fixtures (stoves, cutting boards, plates) cheaply and avoids per-frame
motion noise on genuinely immovable objects.

\paragraph{Per-frame motion gate.}
For non-globally-static objects we form the centroid displacement
$d_t = \lVert \mathbf{c}_t - \mathbf{c}_{t-1}\rVert_2$ and pass it through
a Schmitt trigger (\secref{app:engine:robust}) with thresholds
$(d_{\text{lo}}, d_{\text{hi}}) = (2,4)$\,px, yielding a hysteresis-stable
binary motion signal $m_t$ that does not flicker around the threshold.

\paragraph{Per-hand grasp signal.}
For each hand $h \in \{L, R\}$ we OR-combine three binary contact
indicators into $g_h^{(t)}$, in order of reliability:
(i)~\emph{2D mask overlap} (primary): the rasterized \mano hand mask and
the object mask overlap by at least $30$\,px;
(ii)~\emph{3D fingertip proximity} (fallback): a fingertip joint lies
within $6$\,cm of the back-projected object point cloud $\mathcal{P}^o_t$;
(iii)~\emph{3D wrist proximity} (fallback): the wrist joint lies within
$5$\,cm of $\mathcal{P}^o_t$. The 2D overlap is primary because it is
immune to monocular-depth noise and to the hand-mask subtraction that
drains the 3D cloud when the hand fully wraps the object; the 3D
fallbacks cover frames where \wilor returns an incomplete mesh.

\paragraph{Temporal smoothing.}
Each per-hand signal is filtered with the morphological close-and-drop
operator of \secref{app:engine:robust}: internal gaps of $\le 30$ frames
($\sim 2$\,s at $15$\,fps) are bridged (handling brief mid-action finger
lifts), and runs shorter than $8$ frames ($\sim 0.5$\,s) are removed. We
write the resulting smoothed per-hand signal $\hat g_h^{(t)}$, which
feeds the hierarchical assignment below.

\begin{figure}[t]
  \centering
  \begin{subfigure}[t]{0.56\linewidth}
    \centering
    \includegraphics[width=\linewidth]{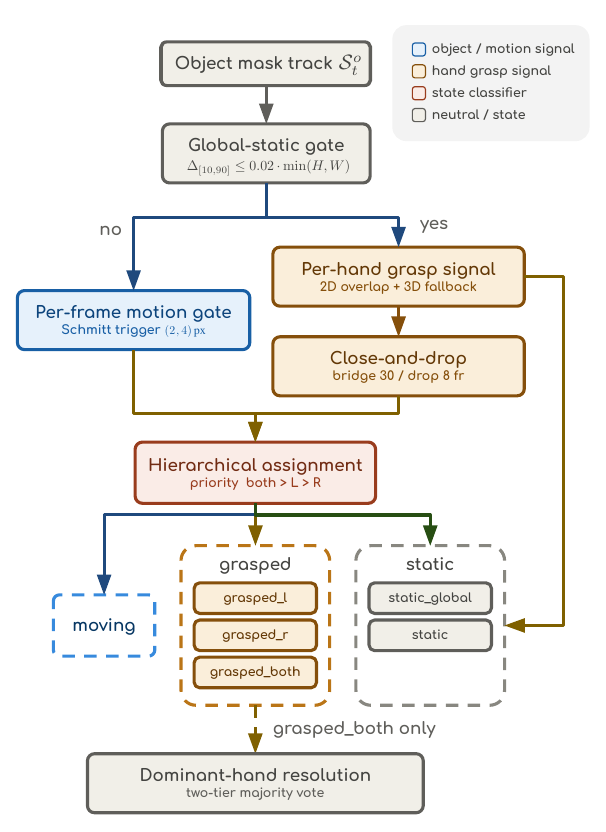}
    \caption{Interaction-state classifier and dominant-hand resolution
    (\secref{app:engine:states}): the global-static gate, the per-frame
    motion gate (Schmitt), the per-hand grasp signal (2D overlap with 3D
    fallbacks) and its morphological smoothing, the hierarchical
    assignment into the six states $\Sigma$, and the two-tier resolution
    branching off \textsc{grasped\_both}.}
    \label{fig:app_state_machine}
  \end{subfigure}\hfill
  \begin{subfigure}[t]{0.41\linewidth}
    \centering
    \includegraphics[width=\linewidth]{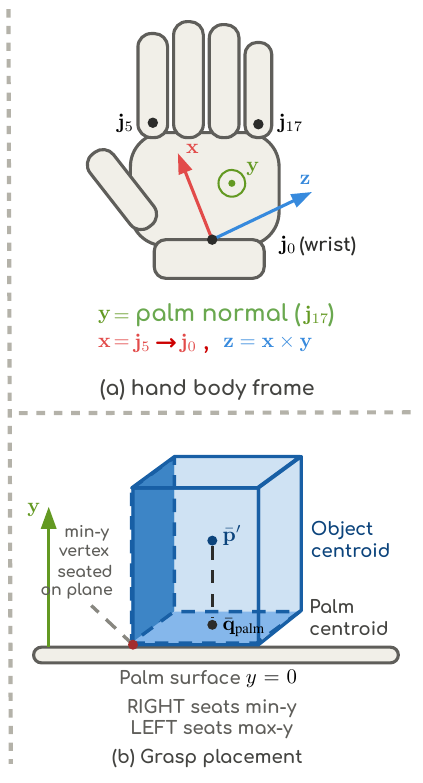}
    \caption{Hand body frame and chirality-aware grasp placement
    (\secref{app:engine:bind}): the palm-landmark construction of
    $({}^{c}\Rmat^h_t, {}^{c}\tvec^h_t)$ from
    $\mathbf{j}_0, \mathbf{j}_5, \mathbf{j}_{17}$ with $+y$ out of the
    palm (left), and a mesh-thickness extreme seated on the $y{=}0$ palm
    surface with the chirality flip (right).}
    \label{fig:app_handbind}
  \end{subfigure}
  \caption{Geometry of the data engine: the per-object interaction-state
  classifier (\subref{fig:app_state_machine}) and the rigid hand bind
  used for \textsc{grasped} segments (\subref{fig:app_handbind}).}
  \label{fig:app_engine_geometry}
\end{figure}

\paragraph{State composition.}
With the gates above, the label is assigned hierarchically (Fig.~\ref{fig:app_state_machine}):
\begin{equation}
  \sigma_t =
  \begin{cases}
    \textsc{static\_global} & \text{globally static} \\
    \textsc{grasped\_both}  & \text{else if } \hat g_L^{(t)} \wedge \hat g_R^{(t)} \\
    \textsc{grasped\_l}     & \text{else if } \hat g_L^{(t)} \\
    \textsc{grasped\_r}     & \text{else if } \hat g_R^{(t)} \\
    \textsc{moving}         & \text{else if } m_t \\
    \textsc{static}         & \text{otherwise.}
  \end{cases}
\end{equation}

\paragraph{Dominant-hand resolution.}
\textsc{grasped\_both} frames are reduced to a single dominant hand for
the rigid bind (\secref{app:engine:bind}) by a two-tier policy.
\emph{Tier 1} uses the clip-level majority of unambiguous single-hand
frames (one side has explicit frames and the other has none, or a $5{:}1$
majority among non-\textsc{both} frames); the whole clip then folds to
that hand. \emph{Tier 2}, used only when no unambiguous single-hand
frame exists, computes per-frame minimum fingertip-to-cloud distance for
each hand and takes the clip-wide majority; minimum fingertip distance is
preferred over wrist proximity because it is robust to elongated tools
whose centroid sits far from the grip.

% --- Combined engine-geometry figure (merged: state machine + hand bind).
% --- Requires \usepackage{subcaption}. If the hand-bind image keeps
% --- internal "(a)/(b)" panel labels, drop them to avoid clashing with
% --- the subfigure (a)/(b) below.

% ---------------------------------------------------------------------
\subsection{State-Dependent Pose Sources}
\label{app:engine:pose}
% ---------------------------------------------------------------------
The per-frame pose ${}^{c}\hat{\pose}^o_t = ({}^{c}\hat{\Rmat}^o_t, {}^{c}\hat{\tvec}^o_t)$
decomposes into a translation and a rotation, each driven by a different
signal depending on $\sigma_t$ (\tabref{tab:pose_sources}). The central
design choice is that the interaction state selects \emph{which} geometric
signal is reliable enough to drive each degree of freedom, rather than
relying on a single state-blind estimator.

\begin{table}[h]
\centering
\small
\setlength{\tabcolsep}{5pt}
\begin{tabular}{lll}
\toprule
State & Translation ${}^{c}\hat{\tvec}^o_t$ & Rotation ${}^{c}\hat{\Rmat}^o_t$ \\
\midrule
\textsc{static\_global} & $\mathrm{med}_t\, \mathbf{c}^{\text{pca}}_t$ (whole clip) & $\mathbf{R}^{\text{cano}}$ (SAM-3D) \\
\textsc{static}         & $\mathrm{med}_{t\in\mathcal{S}}\, \mathbf{c}^{\text{pca}}_t$ (per stretch) & $\mathbf{R}^{\text{cano}}$ (SAM-3D) \\
\textsc{moving}         & mask-bbox-center back-projection (smoothed) & per-frame PCA, sign-corrected \\
\textsc{grasped\_*}     & ${}^{c}\Rmat^h_t\,\mathbf{t}_{\text{canon}} + {}^{c}\tvec^h_t$ & ${}^{c}\Rmat^h_t\,\Rmat_{\text{canon}}$ \\
\bottomrule
\end{tabular}
\vspace{5pt}
\caption{State-dependent pose sources. $\mathbf{c}^{\text{pca}}_t$ is the
robust point-cloud centroid (defined below, using the MAD filter of
\secref{app:engine:robust}); the \textsc{grasped} rows are the rigid hand
bind of \secref{app:engine:bind}.}
\label{tab:pose_sources}
\end{table}

\paragraph{Robust static centroid.}
Both static categories localize translation at the robust centroid of
the back-projected mask cloud $\mathcal{P}^o_t$: after a MAD outlier
filter (\secref{app:engine:robust}) we take the inlier mean
$\mathbf{c}^{\text{pca}}_t$. Locking the mesh to this centroid (rather
than a raw point mean or a bbox center, which are density-biased and
outlier-sensitive respectively) keeps the posed mesh coincident with its
rendered bounding box on every static frame.

\paragraph{Moving translation.}
For \textsc{moving} frames we use the eroded-mask bbox center
$(\min+\max)/2$ of the back-projection, Gaussian-smoothed
($\sigma = 2$ frames) along the trajectory. \textsc{moving} frames are
rare and transient (pickup / setdown), so residual error here has limited
downstream impact.

\paragraph{Per-segment depth realignment.}
Edge blur on hand--object boundaries leaves a small ($1$--$10$\,cm)
systematic depth offset between a grasped object and the hand. Within
each grasp segment we estimate a $z$-correction by matching the posed
object mesh to the hand-vertex surface in the contact region: for each
frame we take the $K{=}20$ nearest (hand-vertex, mesh-point) pairs,
reject the frame if the closest pair exceeds $20$\,cm (residual grasp
false positive), and compute $\Delta z_t = \overline{z_{\text{hand}}} - \overline{z_{\text{mesh}}}$.
A Savitzky--Golay filter (window $7$, order $2$) smooths $\Delta z$ within
the segment, the correction is capped at $\pm 0.1$\,m, and it is applied
jointly to the mesh translation, cached object points, and bounding box
so all surfaces stay consistent. Using the integrated mesh (canonical
scale + robust centroid) as the depth reference, rather than the noisy
per-frame observation, anchors a stable contact plane.

% ---------------------------------------------------------------------
\subsection{Rigid Hand Binding}
\label{app:engine:bind}
% ---------------------------------------------------------------------
This section details the ${}^{c}\hat{\pose}^o_t = {}^{c}\pose^h_t \cdot \mathbf{T}^{\text{cano}}$
rule of \secref{sec:engine:refinement} for \textsc{grasped} segments,
where $\mathbf{T}^{\text{cano}} = (\Rmat_{\text{canon}}, \mathbf{t}_{\text{canon}})$
is the constant hand-relative transform aggregated over a grasp segment.
Note this is distinct from the SAM-3D object orientation
$\mathbf{R}^{\text{cano}}$ (\secref{app:engine:coords}), despite the
similar name.

\paragraph{Hand body frame.}
As illustrated in \figref{fig:app_handbind}, for each frame with a single dominant hand we build an articulation-invariant SE(3) hand frame ${}^{c}\pose^h_t = ({}^{c}\Rmat^h_t, {}^{c}\tvec^h_t)$
from three palm landmarks (wrist $\mathbf{j}_0$, index MCP
$\mathbf{j}_5$, pinky MCP $\mathbf{j}_{17}$):
\begin{equation}
  \mathbf{x} = \frac{\mathbf{j}_5 - \mathbf{j}_0}{\lVert \mathbf{j}_5 - \mathbf{j}_0\rVert}, \quad
  \mathbf{v} = (\mathbf{j}_{17}-\mathbf{j}_0) - \big((\mathbf{j}_{17}-\mathbf{j}_0)^\top\mathbf{x}\big)\mathbf{x}, \quad
  \mathbf{y} = \frac{\mathbf{x}\times\mathbf{v}}{\lVert\mathbf{x}\times\mathbf{v}\rVert}, \quad
  \mathbf{z} = \mathbf{x}\times\mathbf{y},
\end{equation}
with ${}^{c}\Rmat^h_t = [\mathbf{x}\;\mathbf{y}\;\mathbf{z}]$ and
${}^{c}\tvec^h_t = \mathbf{j}_0$. The frame depends only on the palm, not
on finger flexion. The cross-product construction is right-handed for the
RIGHT hand; the resulting $+y$ axis points out of the palm for the RIGHT
hand and into the palm for the LEFT hand, a chirality handled explicitly
in the placement below.

\paragraph{Per-segment canonical pose.}
A grasp segment is a maximal run of frames sharing one
\textsc{grasped\_*} state and one dominant hand; a mid-grasp hand change
splits the segment. With the per-frame relative pose
$\mathbf{T}^{\text{rel}}_t = ({}^{c}\pose^h_t)^{-1}\,{}^{c}\pose^o_t$
(rotation $\Rmat^{\text{rel}}_t = ({}^{c}\Rmat^h_t)^\top {}^{c}\Rmat^o_t$,
translation $\tvec^{\text{rel}}_t = ({}^{c}\Rmat^h_t)^\top({}^{c}\tvec^o_t - {}^{c}\tvec^h_t)$)
computed from the observed pre-refinement pose, we aggregate a constant
canonical pose: $\Rmat_{\text{canon}}$ is the chordal mean on $SO(3)$
(\secref{app:engine:robust}) of $\{\Rmat^{\text{rel}}_t\}$ after rejecting
rotations more than $30^\circ$ from a robust seed (segment-middle frame),
and $\mathbf{t}_{\text{canon}}$ is the per-axis MAD-filtered median of
$\{\tvec^{\text{rel}}_t\}$; this translation is provisional and is
replaced by the geometric placement below, whereas $\Rmat_{\text{canon}}$
is final. The pose is propagated as
${}^{c}\hat{\pose}^o_t = {}^{c}\pose^h_t\,\mathbf{T}^{\text{cano}}$ for
$t \in \mathcal{S}$, making the object rigidly bound to the hand by
construction (zero relative motion).

\paragraph{Geometric placement.}
Because the visible portion of a grasped object is asymmetric about the
held end, the observation-derived $\mathbf{t}_{\text{canon}}$ is biased.
We replace it with a geometric placement from mesh and palm anatomy. Let
$\mathbf{p}'_i = \Rmat_{\text{canon}}\mathbf{p}_i$ be mesh points in the
hand frame with centroid $\bar{\mathbf{p}}'$, and let
$\bar{\mathbf{q}}_{\text{palm}}$ be the centroid of five palm landmarks
$\{\mathbf{j}_0,\mathbf{j}_5,\mathbf{j}_9,\mathbf{j}_{13},\mathbf{j}_{17}\}$.
In-plane components align the centroids,
$\mathbf{t}_{\text{canon}}^{(x,z)} = \bar{\mathbf{q}}_{\text{palm}}^{(x,z)} - \bar{\mathbf{p}}'^{(x,z)}$,
while the out-of-plane component seats a mesh-thickness extreme on the
palm surface ($y=0$):
$\mathbf{t}_{\text{canon}}^{(y)} = -\min_i \mathbf{p}'^{(y)}_i$ for the
RIGHT hand and $-\max_i \mathbf{p}'^{(y)}_i$ for the LEFT (chirality flip
from the body-frame construction). The result flushes the object's
palm-facing face against the palm for both hands.

\paragraph{Boundary smoothing.}
When adjacent segments use different canonical poses (hand change or
re-grip), we remove the discontinuity with SLERP on rotation and linear
interpolation on translation over a $5$-frame ramp, applied both between
two grasp segments and between a grasp and an adjacent non-grasp segment.

% ---------------------------------------------------------------------
\subsection{Robust-Statistics Primitives}
\label{app:engine:robust}
% ---------------------------------------------------------------------
The refinement reuses a small set of robust operators.

\paragraph{MAD outlier filter.}
For points $\{\mathbf{p}_i\}$ with median $\mathbf{m}$, residuals
$d_i = \lVert\mathbf{p}_i-\mathbf{m}\rVert_2$, $\tilde d = \mathrm{med}(d_i)$,
and $\mathrm{MAD} = \mathrm{med}(|d_i - \tilde d|)$, inliers satisfy
$d_i \le \tilde d + 3\cdot 1.4826\cdot\mathrm{MAD}$; the constant $1.4826$
makes MAD a consistent estimator of the standard deviation under
normality. The per-axis variant applies the same test independently per
coordinate and keeps points that are inliers on all axes.

\paragraph{Chordal mean on $SO(3)$.}
For inlier rotations $\{\Rmat^{(k)}\}$,
$\bar{\Rmat} = \mathrm{Proj}_{SO(3)}\!\big(\tfrac{1}{K}\sum_k \Rmat^{(k)}\big)$,
where the projection is the SVD Procrustes map: with
$M = U\Sigma V^\top$, $\bar{\Rmat} = U\,\mathrm{diag}(1,1,\det(UV^\top))\,V^\top$.

\paragraph{Schmitt-trigger hysteresis.}
For a signal $x_t$ with thresholds $(x_{\text{lo}}, x_{\text{hi}})$, the
binary state turns on when $x_t > x_{\text{hi}}$ and off when
$x_t < x_{\text{lo}}$, preventing flicker near a single threshold.

\paragraph{Morphological close-and-drop.}
For a binary sequence with bridge $B$ and minimum run $L$: every
$0$-run of length $\le B$ flanked by $1$s is filled, then every $1$-run
of length $< L$ is removed; edge-bounded runs are not bridged.

% ---------------------------------------------------------------------
\subsection{Perception-Module Configuration}
\label{app:engine:perception}
% ---------------------------------------------------------------------
The off-the-shelf modules of \secref{sec:engine:tracking} are configured
as follows.
\moge (ViT-L, fp16) produces a per-frame metric depth map $D_t$ and a
fitted focal length, used without temporal aggregation.
\flowthreer provides the dense depth used for back-projection;
per-frame depth is temporally aligned against a static-background
template (per-pixel median over flow-classified background pixels) with
the dynamic-mask region excluded.
\geocalib is run on three evenly spaced frames to recover the gravity
vector ${}^{c}\gravity$.
Hand reconstruction runs a YOLO detector for hand boxes and handedness,
\wilor (DINOv2-L backbone, fp16) for the per-detection \mano mesh
($21$ joints, $778$ vertices), and rescales the \mano root translation
into metric units by multi-scale alignment of fingertip projections
against $D_t$; a clip-level handedness vote removes label flips, a
$\sim 35$M-parameter motion infiller fills missing frames, biomechanical
swing--twist limits clamp finger joints, and a Savitzky--Golay filter
(window $9$, order $3$, in quaternion space) removes jitter.
\memfof (fp16) provides per-frame dense optical flow, used only to
classify static-background versus dynamic pixels for the \flowthreer
depth template above; the object pose itself is geometry-driven and
consumes no optical flow.
\samthree detects a region per text prompt on a subset of frames and
initializes a \samtwo streaming track (forward + backward) for the full
clip; a prompt-aware NMS and a containment merge collapse near-duplicate
prompts, keeping up to seven objects. \samthreed runs once per object on
its cleanest unoccluded frame, returning the canonical mesh
$\mathcal{M}^o$, a canonical orientation, and a canonical metric scale.

% ---------------------------------------------------------------------
\subsection{Coordinate Conventions}
\label{app:engine:coords}
% ---------------------------------------------------------------------
All geometry lives in the OpenCV camera-world frame (right-handed, $+x$
right, $+y$ down, $+z$ into the scene), centered at the static camera;
the focal length comes from \moge, so no calibration is required.
\samthreed returns its canonical orientation as a row-form PyTorch3D
quaternion $q$; we convert it to the left-applied column-form rotation
$\mathbf{R}^{\text{cano}}$ used in the main text by
\begin{equation}
  \mathbf{R}^{\text{cano}} = F\,\Rmat_q^{\top}, \qquad F = \mathrm{diag}(-1,-1,+1),
\end{equation}
where $\Rmat_q$ is the matrix of $q$ and $F$ maps the PyTorch3D camera convention to OpenCV's. An object-canonical point $\mathbf{p}_c$ maps to
the camera frame as
${}^{c}\mathbf{p}_t = {}^{c}\Rmat^o_t (s_o \mathbf{p}_c) + {}^{c}\tvec^o_t$
with canonical scale $s_o$.

% ---------------------------------------------------------------------

% ---------------------------------------------------------------------
\subsection{Sanity Filtering}
\label{app:engine:sanity}
% ---------------------------------------------------------------------
The sanity checks mentioned at the end of \secref{sec:engine:refinement}
are two automated tests.
\emph{Scale sanity}: \samthreed's monocular scale $s_o$ is occasionally
off by $2$--$5\times$; if the scaled mesh extent exceeds $1.8\times$ the
mask-implied size $s^{\text{mask}} = \max(W_{\text{mask}}, H_{\text{mask}})\cdot \bar D / f$,
we override $s_o$ with the mask-implied scale.
\emph{Spurious flagging}: objects whose 3D centroid is more than $0.5$\,m
from all hand activity and whose 2D centroid moves less than $10$\,px
across the clip are flagged as background false matches; flagged objects
are dimmed rather than deleted so downstream consumers may keep or drop
them.

% ---------------------------------------------------------------------
\subsection{\actiondataset Annotation Fields and Consolidated Parameters}
\label{app:engine:params}
% ---------------------------------------------------------------------
\paragraph{Semantic prompts.}
The text prompts driving \samthree detection (\secref{sec:engine:tracking})
are taken directly from \actiondataset's hierarchical annotations,
requiring no additional VLM at processing time. We use the short and
detailed action descriptions and the per-clip caption; object nouns are
extracted from these fields (e.g.\ a clip annotated ``slicing a tomato''
yields the prompt ``tomato''), and the same labels are inherited as
language grounding on the output trajectory.

\paragraph{Parameters.}
\tabref{tab:engine_params} lists every threshold in the engine. All are
fixed constants chosen by inspection on a development set, not learned,
and are held constant across the $106$-clip dataset.

\begin{table}[h]
\centering
\small
\setlength{\tabcolsep}{6pt}
\begin{tabular}{lll}
\toprule
Parameter & Value & Use \\
\midrule
Global-static span threshold      & $0.02\cdot\min(H,W)$ & state gate (\secref{app:engine:states}) \\
Motion Schmitt thresholds         & $(2,4)$ px           & state gate \\
Mask-overlap grasp threshold      & $30$ px              & grasp signal \\
Fingertip grasp distance          & $6$ cm               & grasp signal \\
Wrist grasp distance              & $5$ cm               & grasp signal \\
Grasp gap bridge / min run        & $30$ / $8$ frames    & temporal smoothing \\
Rotation outlier threshold        & $30^\circ$           & chordal mean (\secref{app:engine:bind}) \\
MAD inlier multiplier             & $3\cdot 1.4826$      & all robust filters \\
Moving-state Gaussian $\sigma$    & $2$ frames           & moving translation \\
Depth-realign $K$ / far-reject    & $20$ / $0.2$ m       & depth realignment \\
Depth-realign SavGol / cap        & win $7$, ord $2$ / $\pm0.1$ m & depth realignment \\
Boundary SLERP/LERP ramp          & $5$ frames           & segment boundaries \\
\mano SavGol window / order       & $9$ / $3$            & hand smoothing \\
Scale-sanity factor               & $1.8\times$          & sanity (\secref{app:engine:sanity}) \\
Spurious distance / 2D-motion     & $0.5$ m / $10$ px    & sanity \\
\bottomrule
\end{tabular}
\vspace{5pt}
\caption{Consolidated engine parameters. All values are fixed constants,
not learned, and stable across the dataset.}
\label{tab:engine_params}
\end{table}
% =====================================================================
% sections/A2_retarget_details.tex
%
% Appendix B -- Retargeting network architecture & details (SKELETON).
% Expands \secref{sec:retargeting}. Fill the placeholders before the
% camera-ready / supplementary deadline.
%
% Requires \appendix to have been declared in main.tex (see A1 header).
% No new \cite keys are introduced here.
% =====================================================================

% \section{Retargeting Network Details}
\section{Robot Motion Retargeting Details}

\label{app:retarget}
This appendix provides additional implementation details for the cross-embodiment retargeter described in \secref{sec:retargeting}.

\subsection{Network Architecture}
\label{app:retarget:arch}

\begin{figure}[t]
  \centering
  \includegraphics[width=\columnwidth]{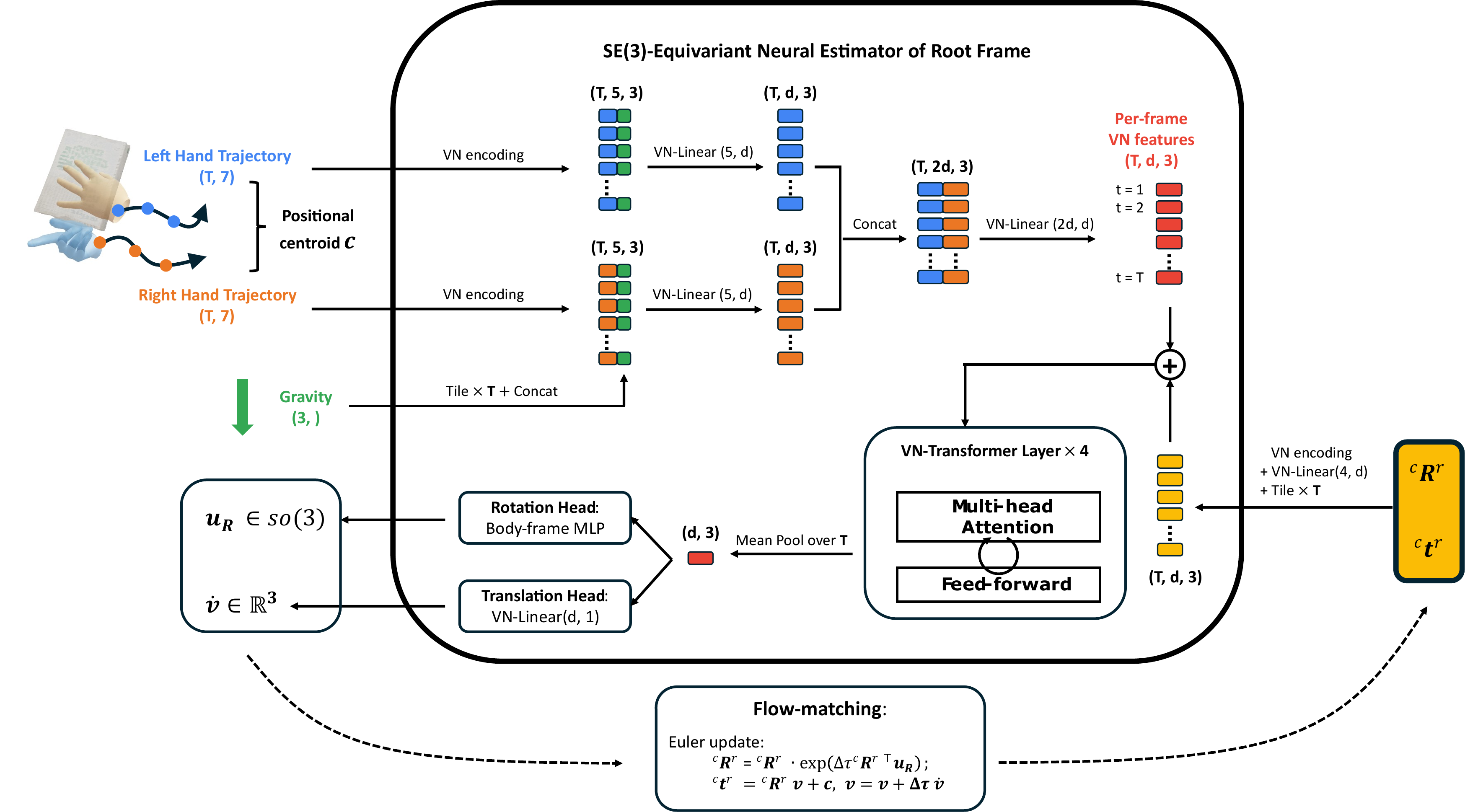}
  \caption{Root-frame estimator architecture.
Bilateral hand trajectories and the optional gravity vector (upper left) are encoded as Vector-Neuron (VN) features and processed by a transformer-based temporal encoder. 
The encoder output is passed to rotation and translation output heads, which predict flow-matching velocities used to denoise a noisy root-frame sample into the final root-frame estimate (yellow, lower right).}
  \label{fig:retarget_appendix}
\end{figure}

\figref{fig:retarget_appendix} illustrates the architecture of the neural root-frame estimator $\Phi$. 
Given bilateral hand trajectories $\{({}^{c}\Rmat^h_t, {}^{c}\tvec^h_t)\in SE(3)\}_{t=1}^T$ and an optional gravity vector ${}^{c}\gravity$, $\Phi$ predicts a distribution over feasible root-frame poses used for downstream IK-based retargeting. 
The network is implemented as a Vector Neuron (VN) flow model: 
all internal geometric features are represented as collections of 3D vectors, preserving SO(3)-equivariance throughout the architecture.

\paragraph{Input representation.}
Each hand trajectory is represented as a tensor of shape $(T,7)$, where $T$ is the trajectory length (i.e., number of time steps) and the 7 dimensions consist of a 3D hand position plus a 4D unit quaternion $(w,x,y,z)$ in the camera frame. 
We compute the bilateral hand centroid $\bm{c} \in\mathbb{R}^3$ by averaging all $2T$ hand positions, and subtract it before VN encoding to remove global translation. 
At each timestep, each hand state is converted into five VN channels:
\begin{equation}
  \left[
  {}^{c}\tvec^h_t-\bm{c},\;
  {}^{c}\Rmat^h_t[:,0],\;
  {}^{c}\Rmat^h_t[:,1],\;
  {}^{c}\Rmat^h_t[:,2],\;
  {}^{c}\gravity
  \right],
\end{equation}
where ${}^{c}\Rmat^h_t \in SO(3)$ is the hand rotation matrix and ${}^{c}\gravity$ is the camera-frame gravity direction, broadcast over time and zeroed when unavailable. 
Thus, each left/right hand trajectory is encoded as VN features of shape $(T,5,3)$, where the five vector channels correspond to the centered hand position, three hand orientation axes, and gravity direction.

\paragraph{VN trajectory encoder.}
Left and right hand trajectory features are first projected independently with a $\mathrm{VN\text{-}Linear}(5,d)$ layer, producing features of shape $(T,d,3)$ for each hand, where $d$ is the channel width (i.e., the number of VN feature channels). 
The two streams are concatenated along the channel dimension and fused with $\mathrm{VN\text{-}Linear}(2d,d)$, yielding a bilateral trajectory feature of shape $(T,d,3)$. 
To condition the flow-matching model, we encode the noisy root-frame state 
$({}^{c}\Rmat^r_{\tau},\,{}^{c}\tvec^r_{\tau}) \in SE(3)$ at flow time $\tau \in [0,1]$ as four VN channels:
\begin{equation}
  \Bigl[
  {}^{c}\tvec^r_{\tau} - \bm{c},\;
  {}^{c}\Rmat^r_{\tau}[:,0],\;
  {}^{c}\Rmat^r_{\tau}[:,1],\;
  {}^{c}\Rmat^r_{\tau}[:,2]
  \Bigr].
\end{equation}
These channels are projected by $\mathrm{VN\text{-}Linear}(4,d)$ to shape $(d,3)$, modulated by a sinusoidal $\tau$-MLP that outputs one scalar scale per channel, broadcast to $(T,d,3)$, and added to the fused bilateral trajectory features. 
This conditioning is injected before the transformer, allowing attention layers to model equivariant interactions between the noisy $SE(3)$ state and trajectory features.
The conditioned sequence of shape $(T,d,3)$ is processed by a VN-Transformer encoder with $L$ blocks. 
Each block applies LayerNorm, multi-head attention with $H$ heads and sinusoidal temporal attention bias, followed by dropout and a feed-forward network of width $d_{\mathrm{ff}}$. 
Mean pooling over time yields a trajectory-level feature of shape $(d,3)$, passed to the output heads.

\paragraph{Output heads.}
Under the flow-matching formulation, the network predicts the velocity field for both rotation and translation. 
For rotation, the pooled VN features of shape $(d,3)$ are rotated into the current body frame using $({}^{c}\Rmat^r_{\tau})^\top$, flattened, passed through an invariant MLP, and mapped back to the camera frame. 
For translation, a VN-Linear head predicts the body-frame offset velocity $\dot{\mathbf{v}}$ from the pooled vector features. 
The root translation is recovered from the predicted offset as
${}^{c}\tvec^r = {}^{c}\Rmat^r \mathbf{v} + \bm{c}$,
where $\bm{c}$ is the bilateral hand-position centroid. 
This parameterization keeps the learned flow field consistent with the equivariant structure of the root-frame estimator. 
At inference, root-frame hypotheses are sampled by integrating the learned model with $N$ Euler steps. 
The full architecture and inference hyperparameters, including VN channel counts, tensor shapes, transformer width/depth, dropout, and flow steps, are summarized in \tabref{tab:arch}.

\begin{table}[t]
\centering
\begin{tabular}{lcc}
\toprule
Parameter & Symbol & Value \\
\midrule
Channel width               & $d$           & 128 \\
Attention heads                & $H$           & 4 \\
Transformer layers             & $L$           & 4 \\
FFN hidden width               & $d_{\mathrm{ff}}$ & 512 \\
Dropout                        & --            & 0.1 \\
Input VN channels per hand     & --            & 5 \\
Conditioning VN channels       & --            & 4 \\
Input shape per hand           & --            & $(T,7)$ \\
VN feature shape per hand      & --            & $(T,5,3)$ \\
Fused feature shape            & --            & $(T,d,3)$ \\
Pooled output shape            & --            & $(d,3)$ \\
Flow ODE steps at inference    & $N$           & 20 \\
\bottomrule
\end{tabular}
\vspace{5pt}
\caption{Model hyperparameters for the root-frame estimator.}
\label{tab:arch}
\end{table}

\subsection{Training Configuration}
\label{app:retarget:training}

All models are trained on a single NVIDIA GeForce RTX 3060 with 12\,GB VRAM. 
Each robot model is trained independently for 500 epochs, with 20 gradient steps per epoch. 
At each step, we draw a fresh batch of 1024 simulated trajectories from the JAX-accelerated physics environment. 
Training takes approximately 1.5--2 hours per robot. 
We use Adam with a fixed learning rate of $10^{-3}$ and gradient clipping at norm 1.0. All training hyperparameters are summarized in \tabref{tab:train}.

\begin{figure}[t]
  \centering
  \includegraphics[width=\columnwidth]{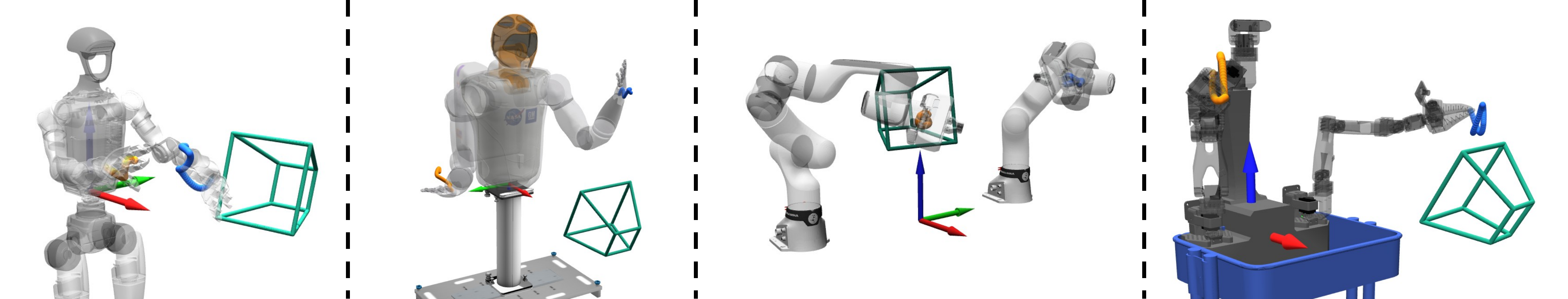}
  \caption{Example sampled training trajectories in simulation for Unitree G1, Robonaut2, dual-Franka, and XLeRobot (left to right). Left-hand trajectories are shown in blue and right-hand trajectories in orange. The robots are rendered semi-transparently for visualization. For each example, the root frame and hand trajectories are projected into a randomly sampled camera frame (green).}
  \label{fig:retarget_train}
\end{figure}

\paragraph{Trajectory generation.}
Training trajectories are generated online via task-space sampling. 
For each environment, $N_{\mathrm{ctrl}}=7$ Cartesian control points are sampled using an Ornstein--Uhlenbeck random walk around a forward-kinematics anchor pose, with per-step noise approximately $0.025\,\mathrm{m}$ and spring constant $0.05$. 
The control points are solved to joint angles with IK and then interpolated with a cubic spline to obtain $T=60$ frames at $f=30\,\mathrm{fps}$, corresponding to a 2-second window.
Example simulated trajectories for different robots are shown in \figref{fig:retarget_train}.

\paragraph{Flow-matching objective.}
We train the network with a flow-matching objective over root-frame poses. 
At each training step, a noisy intermediate root-frame state is sampled along a probability path between a prior sample and the ground-truth root frame. 
The prior rotation is sampled uniformly from $SO(3)$, and the prior translation offset is sampled near the bilateral hand centroid. 
The network predicts the velocity that moves this intermediate state toward the ground truth, with an $\ell_2$ loss applied to both rotation and body-frame translation velocities. 
We use equal loss weights for rotation and translation, $w_R=w_t=1.0$.

\paragraph{Data augmentation.}
To simulate realistic egocentric capture conditions, we apply the following augmentations independently per batch. 
\emph{Position noise} adds zero-mean Gaussian noise with $\sigma_p=0.01\,\mathrm{m}$ to all hand positions, simulating hand-pose tracking jitter. 
\emph{Orientation noise} perturbs each hand quaternion with random rotations up to $0.05\,\mathrm{rad}$. 
\emph{Tracking jumps} occur with probability $p_j=0.20$ and randomly displace one arm trajectory by up to $0.15\,\mathrm{m}$, simulating tracking loss and reacquisition. 
\emph{Hand occlusion} zeros one arm's VN features over a contiguous block of frames with probability $p_o=0.20$, improving robustness to single-hand visibility. 
\emph{Gravity noise} perturbs the camera-frame gravity vector ${}^{c}\gravity$ by up to $0.10\,\mathrm{rad}$; with probability $p_g=0.30$, the gravity channel is zeroed entirely so the model can operate without gravity information. 
Finally, \emph{rear-facing camera} augmentation places the synthetic camera behind the subject with probability $p_c=0.15$, increasing viewpoint diversity.

\begin{table}[t]
\centering
\begin{tabular}{lc}
\toprule
Parameter & Value \\
\midrule
GPU                               & NVIDIA RTX 3060 (12\,GB) \\
Training time per robot           & $\approx$ 1.5--2 hours \\
Epochs                            & 500 \\
Steps per epoch                   & 20 \\
Batch size                        & 1024 \\
Optimizer                         & Adam \\
Learning rate                     & $10^{-3}$ \\
Gradient clip norm                & 1.0 \\
Trajectory length $T$             & 60 frames \\
Frame rate $f$                    & 30\,fps \\
Control points $N_{\mathrm{ctrl}}$ & 7 \\
OU step noise                     & $0.025\,\mathrm{m}$ \\
OU spring constant                & 0.05 \\
Loss weights $(w_R,w_t)$          & $(1.0,1.0)$ \\
\midrule
Position noise $\sigma_p$         & $0.01\,\mathrm{m}$ \\
Orientation noise max             & $0.05\,\mathrm{rad}$ \\
Tracking jump probability $p_j$   & 0.20 \\
Tracking jump magnitude max       & $0.15\,\mathrm{m}$ \\
Occlusion probability $p_o$       & 0.20 \\
Gravity noise max                 & $0.10\,\mathrm{rad}$ \\
Gravity dropout $p_g$             & 0.30 \\
Rear-camera probability $p_c$     & 0.15 \\
\bottomrule
\end{tabular}
\vspace{5pt}
\caption{Training hyperparameters for the root-frame estimator.}
\label{tab:train}
\end{table}

\subsection{Sliding-Window Inference and Optimization}
\label{app:retarget:inference}

Given a video clip with WiLOR-estimated hand trajectories in the camera frame, we first evaluate the root-frame estimator over centered sliding windows. 
The window stride can be adjusted depending on the desired trade-off between temporal resolution and computation. 
Each window produces an $SE(3)$ root-frame estimate by the learned estimator, allowing the prediction to capture gradual body or camera motion rather than relying only on sparse keyframes. 
The estimates are then clustered into $K=5$ candidates using $k$-means under an $SE(3)$ geodesic metric. 
Each candidate is scored by running batched IK over the full trajectory using that candidate as a static reference root frame, and the candidate with the highest bilateral IK convergence rate is selected as the anchor $({}^{c}\Rmat^{r,*},{}^{c}\tvec^{r,*})$.

We then blend each per-frame estimate toward the anchor using separate factors for translation and rotation, with $\alpha_t=0.3$ and $\alpha_r=0.7$. 
The anchor provides global stability, while the per-frame estimates preserve slow root-frame variation over time. 
A Gaussian filter with $\sigma=10$ frames is applied to the blended root trajectory to reduce residual frame-to-frame noise. 
Each camera-frame hand pose is projected into the moving root frame using the corresponding smoothed per-frame $SE(3)$ estimate, yielding root-frame IK targets for every timestep. 
The IK objective tracks these hand targets while using null-space regularization to improve manipulability, avoid self-collision, respect joint limits, and penalize deviation from the robot's default posture. 
Failed frames are filled by linear interpolation from neighboring converged solutions. 
% Finally, the resulting joint trajectories are smoothed with a Gaussian filter of $\sigma=0.1\,\mathrm{s}$.

% =====================================================================
% sections/A3_additional_results.tex
%
% Appendix C -- Additional qualitative results.
% Expands \secref{sec:experiments}.
%
% Cross-references the ego-synthesis subsection \secref{app:engine:ego}
% (Appendix A) and the main-text pipeline figure \figref{fig:pipeline}.
% Figure files (replace placeholder names as needed):
%   C.1 ego grid             : figures/app_C_grid_17x6.jpg
%   C.2 browser screenshot   : figures/app_C_browser.jpg
%   C.3 full pipeline gallery: figures/app_C_pipeline_10clips.jpg
%   C.4 dual-Franka skills   : figures/app_C_franka_skills.jpg
%   C.5 LEAP grasping        : figures/app_C_leap_grasp.jpg
% No new \cite keys are introduced here.

\section{Additional Results}
\label{app:results}
This appendix provides additional examples referenced in
\secref{sec:experiments}, organized from data (the curated egocentric
reconstructions and the interactive browser) through the full
video-to-robot pipeline (simulation and real-robot retargeting) to
downstream policy use.

% ---------------------------------------------------------------------
% ---------------------------------------------------------------------
\subsection{Curated \actiondataset Egocentric Reconstructions}
\label{app:results:dataset}
% ---------------------------------------------------------------------
\figref{fig:app_dataset} (Appendix A) shows engine outputs across the
curated subset of \secref{sec:experiments:dataset}: each clip is given as
the original exocentric source frame paired with its synthesized
egocentric re-rendering (hand meshes, object meshes and point clouds, and
6-DoF poses), produced by the exo-to-ego reframing of
\secref{app:engine:ego}. Source videos with no common capture viewpoint
are all reframed into a consistent, embodiment-aligned egocentric
observation, which is the view exposed by the browser's exo/ego toggle
(\secref{app:results:browser}) and the sense in which \egoinf produces
egocentric data at corpus scale.

The same gallery also illustrates the engine's operating envelope. Rather
than devoting a separate figure to failures, we note the representative
modes visible in \figref{fig:app_dataset}: under severe hand occlusion or
when the hand fully wraps the object, the object cloud is partially
drained and the recovered mesh can be incomplete; reflective or
transparent surfaces yield noisy depth and unstable point clouds; an
imperfect \samthreed mesh or a \samthree mis-segmentation of the target
object occasionally produces a wrong or coarse object geometry. These
cases are consistent with the limitations discussed in
\secref{sec:limitations}; the sanity checks of \secref{app:engine:sanity}
suppress or flag the most implausible ones rather than deleting them.

\begin{figure*}[t]
  \centering
  \includegraphics[width=\linewidth]{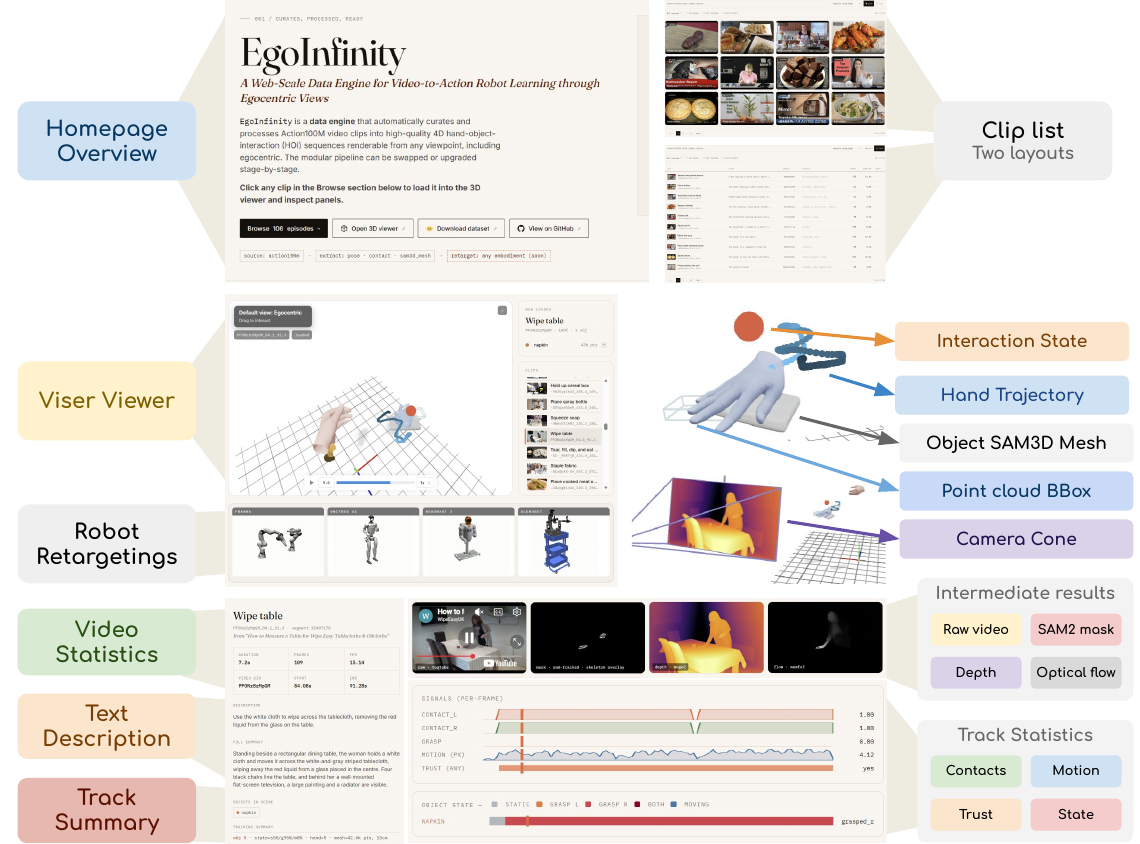}
  \caption{The interactive dataset browser. A static Viser client served
  with no runtime backend lets a reader browse and inspect the curated
  subset. \emph{Homepage and clip list:} a searchable gallery of all
  episodes in grid or table layout. \emph{Viser viewer:} the interactive
  3D scene exposing the hand trajectory, object \samthreed mesh, point
  cloud and bounding box, interaction state, and camera cone, with a
  robot-retargeting panel. \emph{Side panels:} video statistics, the
  \actiondataset text description, per-frame intermediate results (raw
  video, \samtwo mask, depth, optical flow), and a track summary with
  per-frame contact / motion / trust signals and the object interaction
  state (\secref{app:results:browser}).}
  \label{fig:app_browser}
\end{figure*}

% ---------------------------------------------------------------------
\subsection{Interactive Dataset Browser}
\label{app:results:browser}
% ---------------------------------------------------------------------
The full curated subset is also released as an interactive web viewer,
shown only in miniature in the main-text system figure
(\figref{fig:pipeline}) and enlarged in \figref{fig:app_browser}. Each
clip's reconstructed scene is serialized offline with Viser
(\texttt{get\_scene\_serializer()}) and compiled to a self-contained
client (\texttt{viser-build-client}), so the page runs entirely in the
browser with no runtime backend and the same assets back both the static
figures of \secref{app:results:dataset} and the interactive views. The
hosting URL is withheld for anonymity and provided in the camera-ready
version.

For each clip the browser exposes the same quantities as the offline
engine output, all scrubbable along a synchronized timeline. The Viser
viewer renders the metric \mano hand trajectory, the posed object
\samthreed mesh, the point cloud and 3D bounding box, the recovered
interaction state, and the camera cone, with a viewpoint toggle between
the original exocentric camera and the synthesized egocentric view
(\secref{app:engine:ego}). Side panels surface the per-frame intermediate
results (raw video, \samtwo mask, depth, and optical flow), the
\actiondataset text description, video-level statistics, and a track
summary with per-frame contact, motion, and trust signals together with
the object interaction state. The same clip can also be compiled onto the
supported robot embodiments directly in the viewer.

% ---------------------------------------------------------------------
\subsection{Video-to-Robot Pipeline Gallery}
\label{app:results:retarget}
\begin{figure*}[t]
  \centering
  \includegraphics[width=\linewidth]{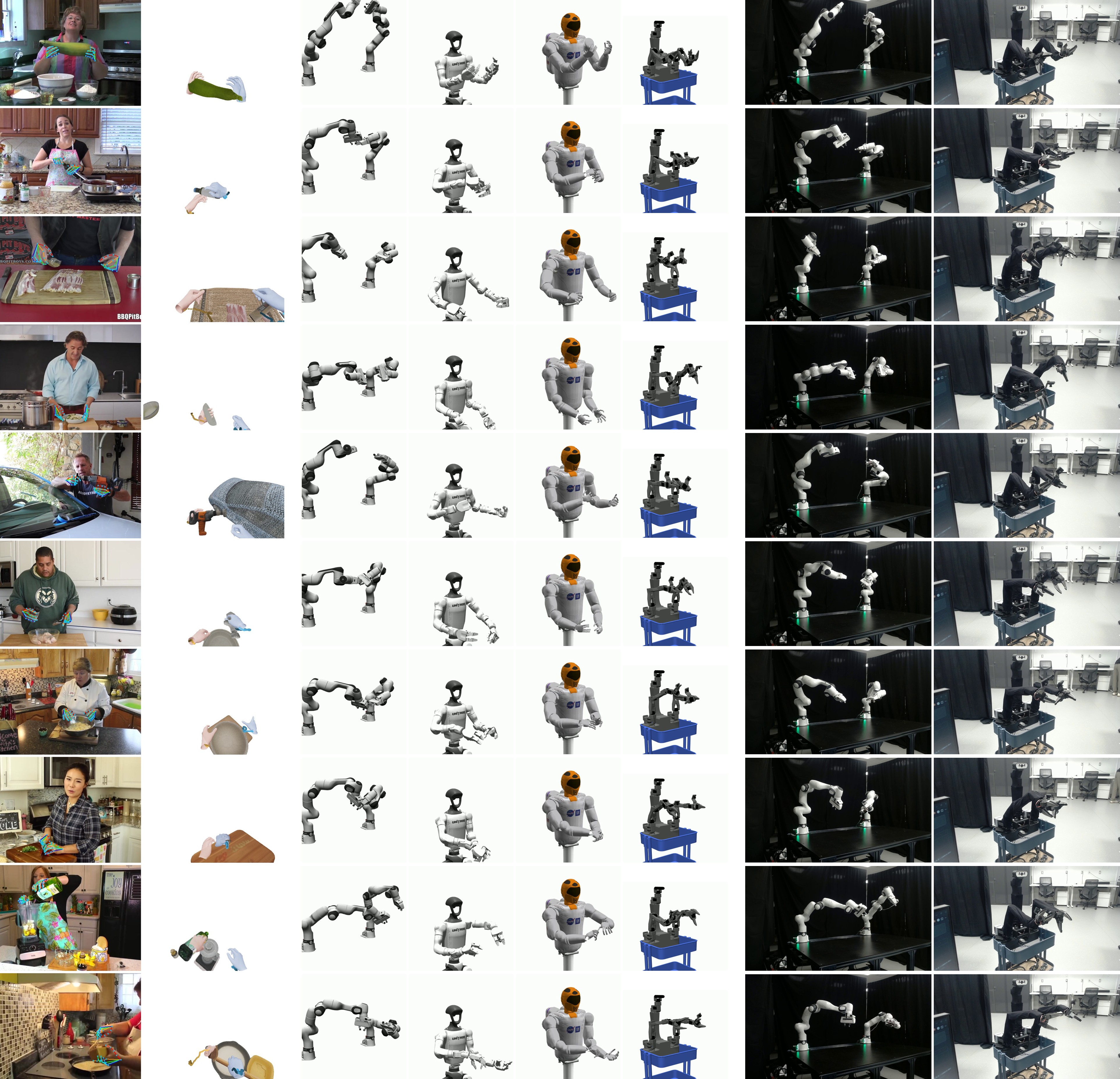}
  \caption{End-to-end video-to-robot pipeline on ten sampled clips. From
  left: raw exocentric frame, reconstructed egocentric view, simulation
  retargeting on dual-arm Franka~FR3, Unitree~G1, Robonaut2, and XLeRobot, and real-robot execution. One recovered trajectory compiles onto multiple embodiments and transfers from simulation to hardware.}
  \label{fig:app_pipeline}
\end{figure*}

% ---------------------------------------------------------------------
\figref{fig:app_pipeline} traces the complete video-to-action pipeline of
\secref{sec:experiments:retargeting} end to end on ten clips sampled from
the curated subset. Each row reads left to right: the raw exocentric
source frame, the reconstructed egocentric hand-object view, the motion
retargeted in simulation onto the dual-arm Franka~FR3 setup, Unitree~G1,
Robonaut2, and XLeRobot, and finally the same motion executed on the two
real-robot platforms. A single recovered 4D trajectory thus compiles onto
substantially different embodiments and transfers from simulation to
hardware, illustrating the agent-agnostic nature of the engine output.

% ---------------------------------------------------------------------
\subsection{Real-Robot Skill Execution}
\label{app:results:real}
\begin{figure*}[t]
  \centering
  \includegraphics[width=\linewidth]{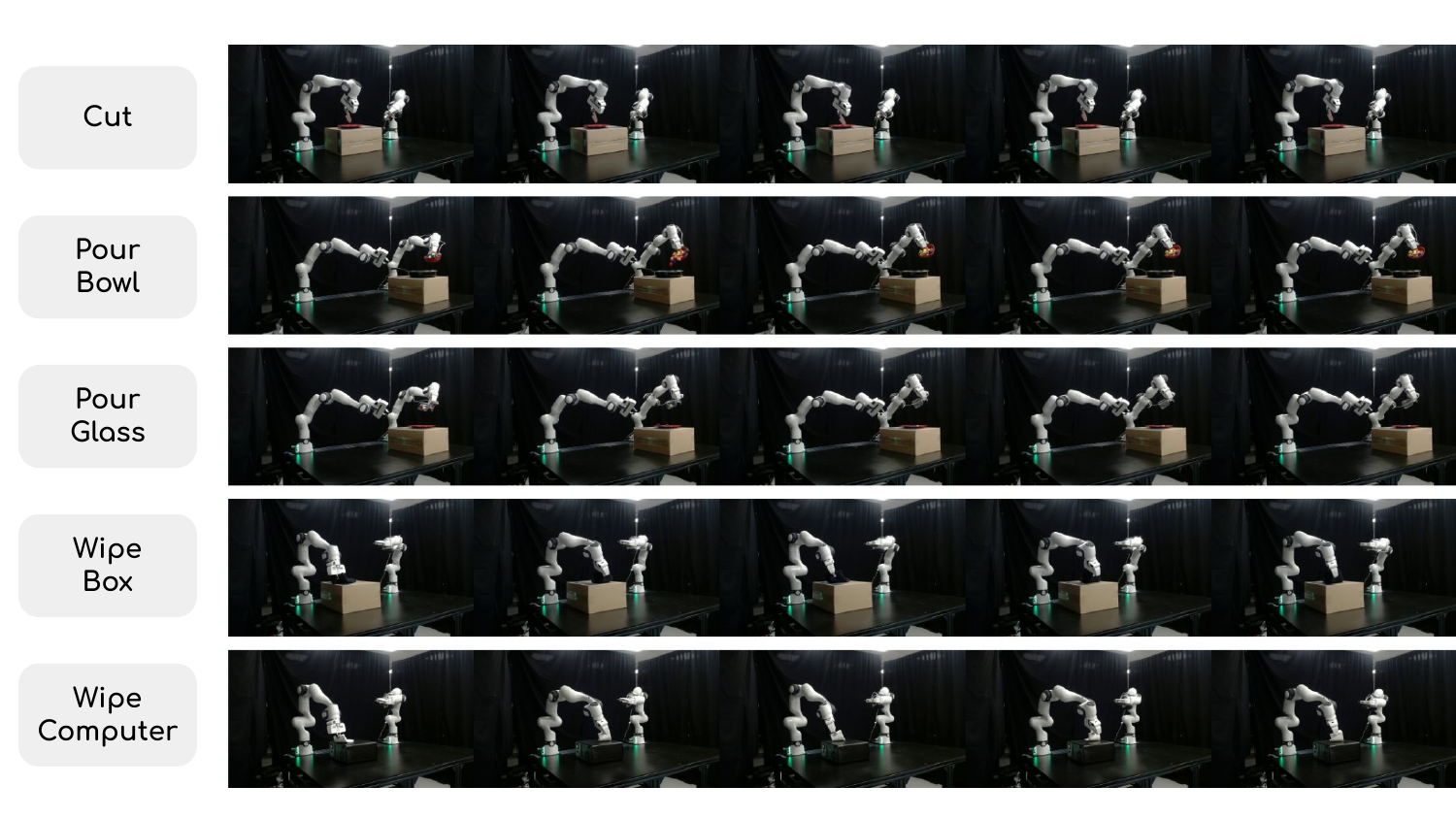}
  \caption{Real-robot skill execution on the dual-arm Franka~FR3. Each
  row is a time-ordered filmstrip of a retargeted skill: Cut, Pour Bowl,
  Pour Glass, Wipe Box, and Wipe Computer.}
  \label{fig:app_franka}
\end{figure*}

% ---------------------------------------------------------------------
\figref{fig:app_franka} shows additional dual-arm Franka~FR3 executions
extending \secref{sec:experiments:real_robot}: retargeted motions
directly replayed on hardware for Cut, Pour Bowl, Pour Glass, Wipe Box,
and Wipe Computer. Each row is a time-ordered filmstrip of one skill,
demonstrating that motions recovered from in-the-wild video drive
functional execution across several distinct manipulation tasks.

% ---------------------------------------------------------------------
\subsection{Downstream Grasping Policy}
\label{app:results:policy}
\begin{figure*}[t]
  \centering
  \includegraphics[width=\linewidth]{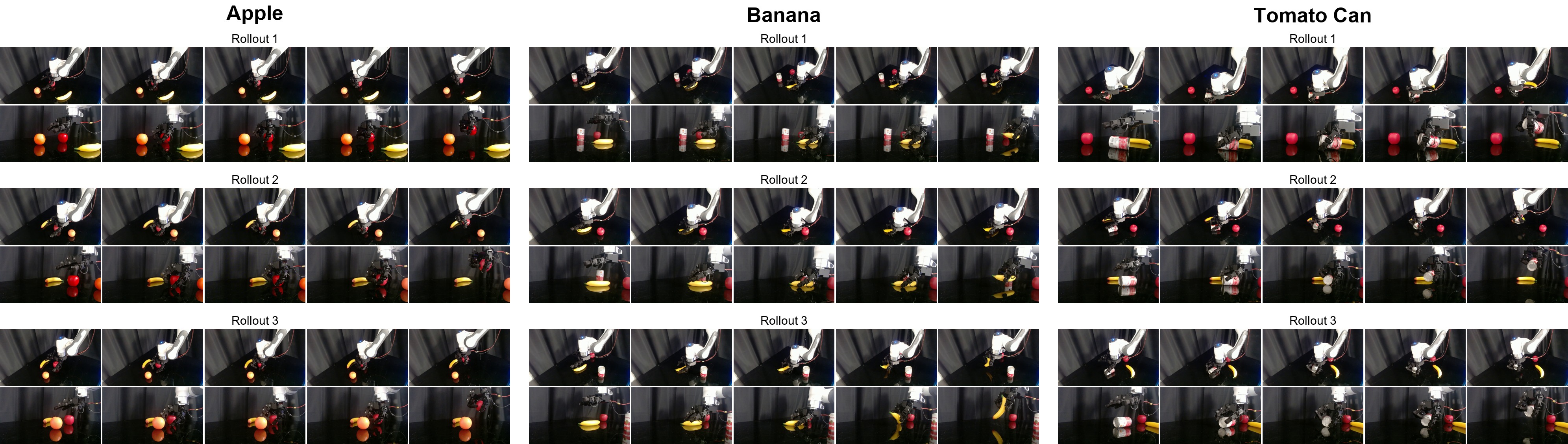}
  \caption{Downstream grasping policy on a real LEAP hand, trained with
  \egoinf-extracted hand motions as priors. Three rollouts each are shown
  for an apple, a banana, and a tomato can, generalizing across object
  shape.}
  \label{fig:app_leap}
\end{figure*}

% ---------------------------------------------------------------------
Beyond direct replay, \figref{fig:app_leap} shows rollouts of a grasping
policy trained on a real LEAP dexterous hand using \egoinf-extracted hand
motions as priors (\secref{sec:experiments:real_robot}). We show three
rollouts each for grasping an apple, a banana, and a tomato can,
generalizing across object shape and demonstrating that the recovered
motions support learned policies rather than open-loop replay alone.

\end{document}